\documentclass[letterpaper,journal]{IEEEtran}

\usepackage{amsmath,amsfonts,amssymb}
\usepackage{algorithm}
\usepackage{algorithmic}
\usepackage{array}
\usepackage[caption=false,font=normalsize,labelfont=sf,textfont=sf]{subfig}
\usepackage{textcomp}
\usepackage{stfloats}
\usepackage{url}
\usepackage{verbatim}
\usepackage{graphicx}
\usepackage{cite}
\usepackage{tikz}
\usepackage{multirow}
\usepackage{amsthm} 

\usetikzlibrary{shapes, arrows.meta, positioning, fit, backgrounds, calc}

\usepackage[utf8]{inputenc}
\usepackage{booktabs}

\DeclareUnicodeCharacter{2265}{$\geq$}
\DeclareUnicodeCharacter{2264}{$\leq$}

\begin{document}

\title{Latent Geometric Chords for Query-Efficient Decision-Based Adversarial Attacks}

\author{Ei Hmue Khine, Yao Li, Jiebao Sun, Shengzhu Shi, Zhichang Guo, and~Boying Wu%
\thanks{This work was supported in part by the National Natural Science Foundation of China under Grant 12401557, Grant 12371419, Grant 12171123, Grant 12271130, and Grant U21B2075; and in part by the National Key R\&D Program of China under Grant 2023YFC2205900 and Grant 2023YFC2205903. (Corresponding author: Yao Li.)}%
\thanks{The authors are with the School of Mathematics, Harbin Institute of Technology, Harbin 150001, China (e-mail: yaoli0508@hit.edu.cn (Y. Li).)}

\thanks{This work has been submitted to the IEEE for possible publication. Copyright may be transferred without notice, after which this version may no longer be accessible.}}

\markboth{Latent Geometric Chords for Query-Efficient Decision-Based Adversarial Attacks}%
{Shell \MakeLowercase{\textit{et al.}}: Bare Demo of IEEEtran.cls for IEEE Journals}
\maketitle

\begin{abstract}
While decision-based black-box adversarial attacks present a severe security threat, current methodologies suffer from fundamental limitations. Pixel-wise attacks frequently introduce unnatural, high-frequency visual artifacts, while latent-space frameworks are confined by the limited search space of low-dimensional manifolds and inherent reconstruction flaws. To resolve these limitations, we propose Latent Geometric Chords (LGC) for Query-Efficient Decision-Based Adversarial Attacks alongside an variant, LGC-H. At its core, LGC navigates decision boundaries by executing a curvature-aware geometric search within a compressed semantic manifold. To guarantee high visual fidelity and circumvent dimensionality bottlenecks, we introduce a Residual-based Adversarial Generation (RAG) mechanism. RAG isolates semantic perturbations as geometric chords and superimposes them directly onto the original source image. RAG substantially resolves baseline reconstruction flaws and effectively doubles the permissible search space dimensions. Experimental results demonstrate that LGC achieves robust cross-dataset transferability and substantially outperforms state-of-the-art baselines. Notably, our method, LGC, minimizes perturbation magnitudes while achieving state-of-the-art visual fidelity---with a Structural Similarity Index Measure (SSIM) exceeding 0.99 and a Learned Perceptual Image Patch Similarity (LPIPS) below 0.01 at 5000 queries---and sustaining high attack success rates under stringent perceptual constraints, successfully compromising adversarially trained robust models. The source code is available at: https://github.com/eihmuekhine/Latent-Geometric-Chords.
\end{abstract}

\begin{IEEEkeywords}
Adversarial Machine Learning (AML), Decision-Based Black-Box Attacks, Latent Space Optimization, Query Efficiency, Semantic Perturbations.
\end{IEEEkeywords}

\section{Introduction}
\IEEEPARstart{M}{odern} visual applications rely heavily on deep neural networks (DNNs) for complex tasks including image classification \cite{dosovitskiy2020image,he2016deep}. Despite their empirical success, DNNs remain highly susceptible to adversarial examples—carefully crafted, visually imperceptible perturbations that deceive models into rendering incorrect predictions \cite{szegedy2014intriguing,goodfellow2015explaining,carlini2017towards}. Based on adversary knowledge, attacks are categorized into white-box and black-box settings. White-box methods require full access to the target model's internal architecture and gradients, a condition rarely satisfied on real-world deployments \cite{carlini2017towards}. Consequently, black-box attacks present a more realistic threat paradigm and are classified into transfer-based, score-based, and decision-based approaches. Transfer-based attacks craft perturbations on a known substitute model to exploit cross-architecture vulnerability, yet success is not guaranteed due to unreliable transferability \cite{papernot2016transferability,wu2020skip}. Score-based attacks iteratively refine perturbations by querying the target model's continuous output probabilities \cite{chen2017zoo,ilyas2018prior}, which is often impractical as commercial APIs typically return only discrete labels. Therefore, decision-based (hard-label) attacks, which optimize perturbations relying exclusively on discrete top-1 class predictions, represent the most profound and practical security threat \cite{brendel2018decision,chen2020hopskipjumpattack,maho2021surfree,wang2022triangle}.

Without gradients or continuous scores, hard-label attackers are completely blind to both the distance and direction of the optimal adversarial path. In a vast, high-dimensional space, they must use inefficient, blind trial-and-error queries to find the decision boundary, requiring thousands of attempts to estimate the available path. To mitigate this excessive query complexity, modern methods introduce geometric acceleration techniques—such as normal vector estimation in HSJA \cite{chen2020hopskipjumpattack} and GeoDA \cite{rahmati2020geoda}, and curvature-aware search in CGBA \cite{reza2023cgba}—to guide adversarial trajectories along decision boundaries. Despite these geometric advancements, existing decision-based attacks optimize perturbations directly in the raw pixel domain. In pixel-wise attacks, adversaries systematically alter individual pixel values to manipulate the model's prediction. Although restricted by rigid $\ell_p$-norm constraints to remain mathematically imperceptible, modifying images at pixel level misaligns with human visual perception and the natural data manifold \cite{chen2023advdiffuser,chen2025diffusion}. These spatial domain manipulations introduce unnatural, semantically meaningless artifacts that deviate from real-world distributions \cite{xue2023diffusion} and primarily exploit brittle, high-frequency ``non-robust features'' that are easily neutralized by robust optimization techniques like adversarial training \cite{ilyas2019adversarial,madry2018towards}. Consequently, a transition toward manipulating robust, high-level semantic concepts is critical \cite{inkawhich2019feature}.

To circumvent the limitations of pixel-wise manipulations, recent studies have explored decision-based latent frameworks like Latent-HSJA \cite{na2022unrestricted}, operating within compressed generative representations. While these approaches improve semantic realism, applying them directly to decision-based black-box settings exposes severe optimization and fidelity drawbacks. First, because the latent dimension $k$ is much smaller than the pixel dimension $n$ ($k \ll n$), confining the adversarial search to this strictly low-dimensional generative manifold drastically limits the available directions for adversarial movement \cite{jalal2017robust, samangouei2018defense, arjovsky2017towards, falconer2014fractal}. Consequently, decision-based latent frameworks frequently stall and waste tens of thousands of queries blindly searching for rare boundary intersections \cite{jang2020topology}. Second, inherent image inversion process inevitably discards high-frequency textures and alters core concepts, producing blurry artifacts and unintended semantic drift that render the resulting images unnatural \cite{xia2023gan}.

Therefore, this paper addresses the following fundamental research question:
    \textit{How can we systematically achieve high query efficiency and eliminate the high-frequency, unnatural visual artifacts prevalent in pixel-wise attacks, while simultaneously overcoming the low-dimensional manifold constraints and reconstruction errors inherent to latent-space adversarial frameworks?}
Addressing this challenge is of critical importance. It provide the theoretical and practical foundation necessary to achieve the precise geometric query-efficiency of pixel-space optimization while preserving the high visual fidelity of unrestricted adversarial attack methods in strict decision-based black-box scenarios.

To address this challenge, we propose Latent Geometric Chords for Query-Efficient Decision-Based Adversarial Attacks (LGC) alongside a highly efficient variant (LGC-H). LGC translates curvature-aware optimization into a compressed semantic manifold. Crucially, we introduce a novel Residual-based Adversarial Generation (RAG) mechanism that isolates semantic shifts as ``geometric chords'' and superimposes them directly onto the pristine input. This strategy conceptually expands the search space permissible up to dimensions $2k$ (as formally proven in Section IV-B), mitigating the dimensionality bottleneck to ensure rapid convergence and maximize query efficiency. Similarly, in RAG, anchoring the perturbation to the original image largely mitigates baseline decoder reconstruction errors, enabling high visual fidelity.

Our primary contributions are summarized as follows:
\begin{itemize}
    \item We propose LGC and LGC-H, translating decision-based black-box adversarial optimization into a compressed semantic manifold. By employing a curvature-guided semicircular search, LGC efficiently navigate highly non-linear decision boundaries to achieve highly competitive query efficiency.
    \item  We introduce the RAG mechanism, isolating semantic perturbations as latent ``geometric chords.'' RAG expands the search space up to $2k$ dimensions, while largely mitigating baseline decoder reconstruction errors by anchoring the semantic shift directly to the pristine input.
    \item Extensive evaluations demonstrate that LGC achieves high visual quality (SSIM $> 0.99$, LPIPS $< 0.01$) with minimal query budgets. In targeted attacks on ResNet50, our method reduces perturbation magnitude by a factor of six compared to state-of-the-art approaches while preserving near-perfect structural similarity. Furthermore, our framework, LGC, exhibits strong cross-dataset generalizability—attacking out-of-distribution datasets such as Places365 and CelebAMask-HQ using a single ImageNet-trained autoencoder—and effectively compromises adversarially trained robust models.
\end{itemize}

\section{Related Work}
\subsection{Decision-Based Geometric Attacks}
Decision-based adversarial attacks operate under highly restrictive threat models, requiring adversaries to optimize perturbations relying exclusively on top-1 predicted labels. Existing decision-based attacks can be broadly divided into two categories: random search attacks and normal-vector-based attacks. Random search frameworks, such as the Boundary Attack \cite{brendel2018decision}, employ rejection sampling to find progressively smaller perturbations along the decision boundary, whereas AHA \cite{li2021aha} generates random samples from a normal distribution biased by the mean of historical queries. RayS \cite{chen2020rays} reformulates the attack as a discrete problem of finding the closest boundary, employing a progressive subdivision strategy that iteratively refines blocks of perturbation to enhance search efficiency, Sign-OPT \cite{chen2020sign} proposes a highly query-efficient approach by computing the sign of the directional derivatives for gradient estimation and Triangle Attack \cite{wang2022triangle} iteratively construct a triangle in a subspace, formed by the spatial relationship between the original benign sample and two consecutive adversarial examples. Similarly, SurFree \cite{maho2021surfree} explores diverse search directions guided entirely by the geometric characteristics of the classifier's decision boundary.

Conversely, normal-vector-based attacks leverage the local normal vector at the decision boundary to systematically guide the perturbation search. HSJA \cite{chen2020hopskipjumpattack} approximates the gradient direction at the decision boundary point via Monte Carlo sampling. Based on this, QEBA \cite{li2020qeba} further estimate the gradient within subspaces in spatial, frequency and intrinsic dimensions. By exploiting the observation that decision boundaries typically have low curvature near data samples, qFool \cite{liu2019geometry} and GeoDA \cite{rahmati2020geoda} achieve efficient gradient estimation by locally approximating the boundary as a flat linear hyperplane. However, the decision boundaries of modern deep neural networks are highly non-linear, containing narrow regions with sharp curvature where rigid linear approximations inevitably fail. To successfully navigate these complex topologies, recent methods such as CGBA \cite{reza2023cgba} have transitioned to non-linear search trajectories. CGBA introduces a highly efficient, normal-guided semicircular search that guarantees boundary intersection regardless of local curvature.

Despite these geometric advancements, a fundamental vulnerability persists in pixel-wise attacks. Modifying images directly at the pixel level disregards human visual perception, introducing unnatural, high-frequency artifacts that structurally degrade image quality and render the attacks easily detectable. To overcome this limitation, our proposed LGC framework adopts the query-efficient, curvature-aware search strategy of CGBA but shifts the entire optimization process into a compressed semantic manifold. By utilizing latent geometric chords instead of raw pixel noise, LGC leverages complex boundary curvature to maintain strong query efficiency while avoiding the visual degradation of pixel-wise attacks.

\subsection{Unrestricted Adversarial Attacks}

Unrestricted adversarial attacks manipulate semantic attributes (e.g., color, rotation, or style) rather than adding small norm-bounded perturbations, producing perceptually natural images that fool deep neural networks \cite{poursaeed2019fine, kakizaki2020adversarial}. While these methods successfully deceive classifiers using naturally plausible images \cite{song2018constructing}, extending unrestricted attacks to black-box settings remains challenging.

An existing method \cite{kakizaki2020adversarial} proposed an image-to-image translation approach but requires hundreds of thousands of queries---impractical for real-world applications.

To address these limitations, Latent-HSJA \cite{na2022unrestricted} combines HSJA decision-based attacks \cite{chen2020hopskipjumpattack} with StyleGAN2 \cite{karras2020analyzing} latent space manipulation. This method achieves query-efficient (under 20,000 queries) targeted unrestricted attacks. However, Latent-HSJA suffers from severe dimensionality bottlenecks (constrained to the low-dimensional GAN manifold $k \ll n$) and inherent reconstruction errors from GAN inversion, limiting search efficiency and visual fidelity. Furthermore, its reliance on domain-specific generative models (e.g face-specific GANs) restricts its applicability in cross-dataset scenarios. These weaknesses motivate our Residual-based Adversarial Generation (RAG) mechanism, which expands the search space up to $2k$ dimensions while significantly reducing decoder reconstruction flaws.

\section{Problem Definition}

Let $C: \mathbb{R}^n \to \mathbb{R}^L$ denote a pre-trained $L$-class classifier. In a \textit{decision-based} (hard-label) black-box setting, the adversary aims to craft an adversarial example $\mathbf{x}_{\mathrm{adv}}$ relying solely on the discrete output predictions $\hat{y} = C(\mathbf{x})$. 

In general, this is formulated as an optimization problem in the raw pixel domain. Given a pristine source image $\mathbf{x}_0 \in [0,1]^n$ correctly predicted as its ground-truth label $y_{\mathrm{orig}}$ by the model $C$, the objective is to minimize the magnitude (typically the $L_p$-norm) of a spatial perturbation $\delta\mathbf{x}$ such that the perturbed image $\mathbf{x}_0 + \delta\mathbf{x}$ causes misclassification:
\begin{equation}
\min_{\delta\mathbf{x} \in \mathbb{R}^n} \|\delta\mathbf{x}\|_p \quad \text{s.t.} \quad \phi(\mathbf{x}_0 + \delta\mathbf{x}) = 1,
\end{equation}
where $\phi(\cdot) \in \{-1, 1\}$ is a binary indicator function denoting adversarial success. 

To solve this efficiently, geometric decision-based attacks such as HSJA \cite{chen2020hopskipjumpattack} and CGBA \cite{reza2023cgba} optimize a specific search direction $\hat{\zeta} \in \mathbb{R}^n$. By moving in this direction starting at $\mathbf{x}_0$, the adversary searches for an adversarial image with the minimal possible perturbation. For a queried image $\mathbf{x}_q = \mathbf{x}_0 + \boldsymbol{d}(\hat{\zeta})$, where $\boldsymbol{d}(\hat{\zeta})$ denotes the perturbation vector scaled along direction $\hat{\zeta}$, the objective shifts to minimizing the perturbation magnitude:
\begin{equation}
\hat{\zeta}^* = \arg \min_{\hat{\zeta} \in \mathbb{R}^n} \|\boldsymbol{d}(\hat{\zeta})\|_2 \quad \text{s.t.} \quad \phi(\mathbf{x}_q) = 1.
\end{equation}

Conversely, unrestricted adversarial attacks redefine the distance metric to evaluate semantic transformations (e.g., rotation, hue, or high-level styles) rather than strict pixel-wise constraints. Recent latent-space methods in the decision-based setting, such as Latent-HSJA \cite{na2022unrestricted}, map the input to a lower-dimensional manifold $\mathbb{R}^k$ ($k \ll n$) using an encoder $E$ and a generative decoder $G$. This framework suggests that if the $L_2$ distance between a pristine latent vector $\mathbf{z}_s = E(\mathbf{x}_0)$ and an adversarial latent vector $\mathbf{z}_{\mathrm{adv}}$ is sufficiently small, the resulting synthesized images will remain perceptually indistinguishable to human observers. Mathematically, this shifts the optimization to:
\begin{equation}
\min_{\mathbf{z}_{\mathrm{adv}} \in \mathbb{R}^k} \|\mathbf{z}_{\mathrm{adv}} - \mathbf{z}_s\|_2 \quad \text{s.t.} \quad \phi(G(\mathbf{z}_{\mathrm{adv}})) = 1.
\end{equation}

To synthesize the query efficiency of geometric attacks with the perceptual benefits of semantic attacks, our goal is to find the optimal direction $\hat{\zeta}^* \in \mathbb{R}^k$ that minimizes the magnitude of the latent perturbation. Translating a semantic baseline $\mathbf{z}_s$ along this unit direction yields a candidate adversarial latent vector $\mathbf{z}_{\mathrm{adv}} = \mathbf{z}_s + \boldsymbol{d}(\hat{\zeta})$, where $\boldsymbol{d}(\hat{\zeta}) \in \mathbb{R}^k$ denotes the required latent perturbation. To produce the final adversarial example $\mathbf{x}_{\mathrm{final}}$, the synthesized image $G(\mathbf{z}_{\mathrm{adv}})$ is integrated with the generated semantic baseline $G(\mathbf{z}_s)$ and the pristine image $\mathbf{x}_0$. The overarching optimization is thus formulated as:
\begin{equation}
\hat{\zeta}^* = \arg \min_{\hat{\zeta} \in \mathbb{R}^k} \|\boldsymbol{d}(\hat{\zeta})\|_2 \quad \text{s.t.} \quad \phi(\mathbf{x}_{\mathrm{final}}) = 1, \; \|\hat{\zeta}\|_2 = 1.
\end{equation}

Depending on the model, the indicator function $\phi(\mathbf{x})$ for a \textit{non-targeted attack} is explicitly defined as:
\begin{equation}
\phi(\mathbf{x}) = 
\begin{cases} 
1, & \text{if } C(\mathbf{x}) \neq y_{\mathrm{orig}} \\
-1, & \text{otherwise.}
\end{cases}
\end{equation}
Conversely, for a \textit{targeted attack} directed at a specific target class $l_t$, it is defined as:
\begin{equation}
\phi(\mathbf{x}) = 
\begin{cases} 
1, & \text{if } C(\mathbf{x}) = l_t \\
-1, & \text{otherwise.}
\end{cases}
\end{equation}

\begin{figure}[t]
    \centering
    \includegraphics[width=0.5\textwidth]{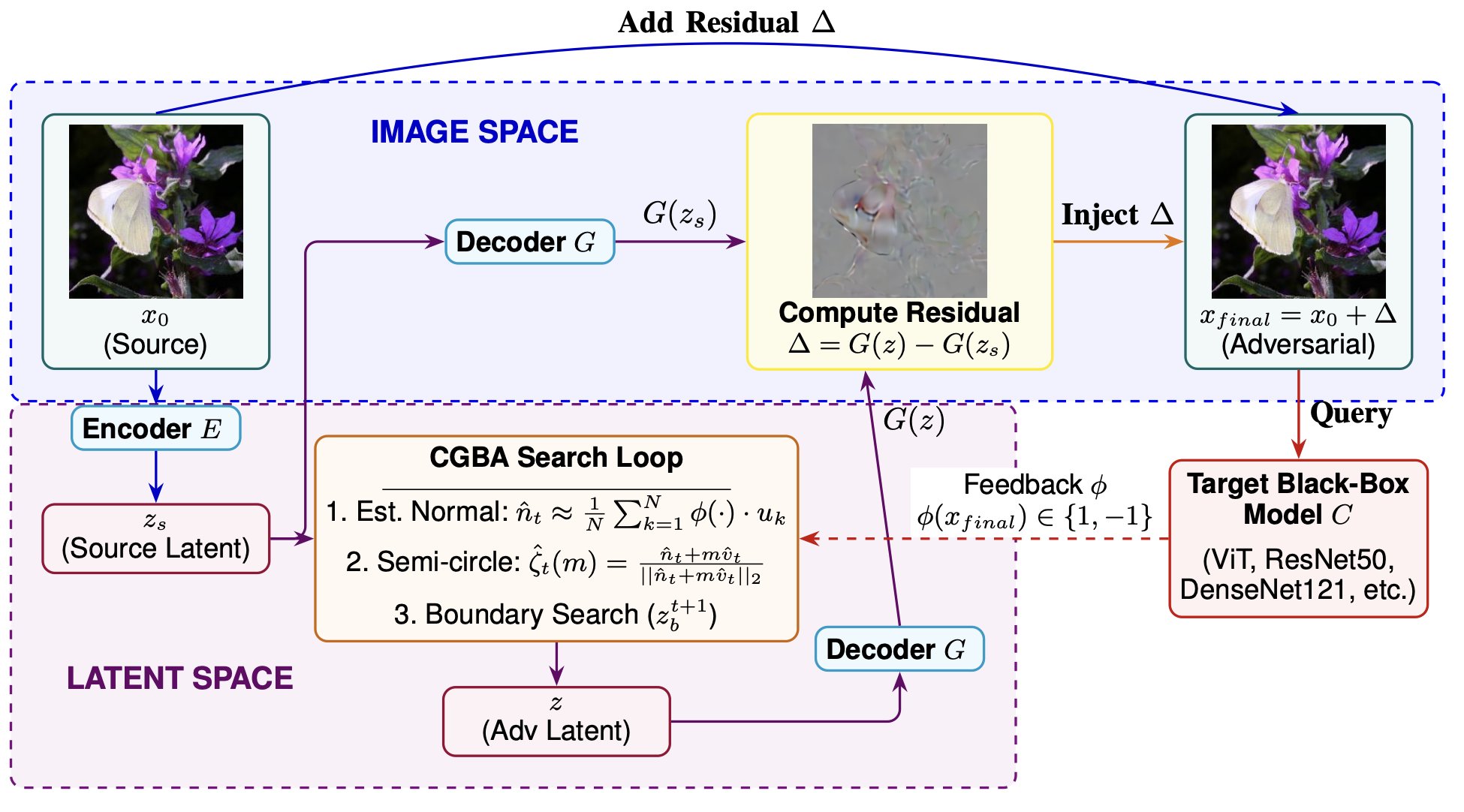}
    \caption{The proposed LGC architecture. The source image $\mathbf{x}_0$ is encoded into a compact latent baseline $\mathbf{z}_s$. Following a curvature-aware geometric optimization (CGBA) in the latent space, the perturbed latent vector is passed through the decoder to generate the high-fidelity adversarial example $\mathbf{x}_{\mathrm{final}}$, which is subsequently queried against target black-box models (e.g., ViT, ResNet-50).}
    \label{fig:architecture}
\end{figure}

\section{Methodology}
This paper presents a decision-based adversarial method conducted entirely within the latent representations of an autoencoder. Unlike conventional attacks (such as the pixel-based Curvature-aware Geometric Black-Box Attack, CGBA) that directly alter raw pixel values and cause visible noise, our proposed Latent Geometric Chords for Query-Efficient Decision-Based Attack modifies semantic features. By operating within the latent manifold, LGC leverages a curvature-guided search algorithm of CGBA to effectively map the decision boundary of the target model. Furthermore, to guarantee visual stealth, LGC incorporates a residual-based technique to isolate the pure semantic shift. The isolated difference between the reconstructions of the altered latent vector and the baseline latent vector is directly superimposed onto the original source image.
This mechanism ensures that the adversarial noise is successfully injected without deteriorating the natural appearance and high-frequency structures of the source image. Moreover, it can expand the adversarial search space up to $2k$ dimensions.

\subsection{Latent Normal Vector Estimation}
Because our threat model assumes strict black-box access, the internal gradient information $\nabla_{\mathbf{x}} C(\mathbf{x})$ of the target classifier is entirely unavailable. Consequently, we must numerically approximate the geometric properties of the decision boundary directly from the latent manifold, adapting numerical approximation techniques from recent literature. Let $\mathbf{z}_b^t$ denote the known boundary intersection at iteration $t$; we employ Monte Carlo sampling---similar to the approach used in the pixel-wise decision-based adversarial attack CGBA---to estimate the local normal vector $\hat{\mathbf{n}}_t$.

By generating $N$ random Gaussian perturbation vectors $\mathbf{u}_k \sim \mathcal{N}(0, \mathbf{I})$ and querying the classifier, the normal direction can be estimated via:
\begin{equation}
\hat{\mathbf{n}}_t \approx \frac{1}{N} \sum_{k=1}^{N} \phi\left(\mathbf{x}_0 + G(\mathbf{z}_b^t + \sigma \mathbf{u}_k) - G(\mathbf{z}_s)\right) \cdot \mathbf{u}_k.
\end{equation}
Here, $\sigma$ defines the sampling step size, and the binary function $\phi(\cdot)$ returns $+1$ if the synthesized query crosses into the adversarial region in the image domain, and $-1$ otherwise. Note that this query image inside the indicator function is constructed using the RAG mechanism (detailed in Section IV-B). This critical approximation process dictates the direction for subsequent boundary exploration within the semantic space.

\subsection{Residual-Based Adversarial Generation (RAG)} 
\label{sec:RAG}

To construct high-fidelity adversarial examples, we address the geometric limitations of standard latent-space optimization by introducing a chord-based perturbation constrained within a $2k$-dimensional space.

\vspace{1.5mm} 
\noindent\textbf{Limitations of Standard Manifold Optimization.} 
Let $G: \mathbb{R}^k \rightarrow \mathbb{R}^n$ (where $k \ll n$) be a generator. Its output forms a non-linear manifold $\mathcal{M} \subset \mathbb{R}^n$ \cite{arjovsky2017towards}: 
\begin{equation} 
\mathcal{M} = \{ G(\mathbf{z}) \mid \mathbf{z} \in \mathbb{R}^k \}. 
\end{equation} 
Standard generative attacks optimize entirely within the latent space, restricting the adversarial sample $\mathbf{x}_{\mathrm{adv}} = G(\mathbf{z}^*)$ to lie strictly on $\mathcal{M}$ \cite{arjovsky2017towards}. Consequently, the search space is bounded by the intrinsic manifold dimension, $\dim_B(\mathcal{M}) \leq k$ \cite{jalal2017robust,samangouei2018defense,arjovsky2017towards, falconer2014fractal}. Due to this dimensionality bottleneck and inherent reconstruction errors \cite{xia2023gan}, $\mathcal{M}$ frequently fails to intersect the target adversarial region within a tight $L_p$-norm ball $\mathcal{B}_\epsilon(\mathbf{x}_0)$ centered around the pristine image $\mathbf{x}_0$.

\vspace{1.5mm} 
\noindent\textbf{Dimensionality Expansion via the Chord Set.} 
To bypass this $k$-dimensional restriction, we mathematically redefine the perturbation mechanism. Given an encoder $E$, a base latent vector $\mathbf{z}_s = E(\mathbf{x}_0)$, and a perturbed latent vector $\mathbf{z}$, we isolate the visual modification as a residual vector $\Delta$. Geometrically, $\Delta$ represents a chord connecting two points on $\mathcal{M}$: $ \Delta = G(\mathbf{z}) - G(\mathbf{z}_s)$ 
The space of all such possible perturbations forms the chord set $\mathcal{C}$: 
\begin{equation} 
\mathcal{C} = \{ G(\mathbf{y}) - G(\mathbf{x}) \mid \mathbf{x}, \mathbf{y} \in \mathbb{R}^k \} = \mathcal{M} + (-\mathcal{M}). 
\end{equation} 
This geometric formulation expands the available perturbation space, as formalized below. Crucially, because the chord set $\mathcal{C}$ is exclusively constructed from the difference of points on the original manifold $\mathcal{M}$, this $2k$-dimensional expansion is not an arbitrary unconstrained space (unlike the raw pixel space $\mathbb{R}^n$ or other random dimensional expansions). Instead, it remains strictly coupled to the generative manifold, ensuring that the perturbations within this expanded $2k$-dimensional space retain semantic meaning.

\vspace{1.5mm} 
\noindent\textbf{Theorem 1.} \textit{Assume $G: \mathbb{R}^k \rightarrow \mathbb{R}^n$ is a Lipschitz continuous mapping. For the generated manifold $\mathcal{M} = G(\mathbb{R}^k)$ and its associated chord set $\mathcal{C} = \mathcal{M} + (-\mathcal{M})$, the Hausdorff dimension satisfies $\dim_H(\mathcal{C}) \leq 2k$.}

\vspace{1.5mm} 
\noindent\textit{Proof.} A Lipschitz continuous mapping does not increase the Hausdorff dimension of its domain \cite[Corollary 2.4]{falconer2014fractal}. Therefore, $\dim_H(\mathcal{M}) \leq \dim_H(\mathbb{R}^k) = k$. Furthermore, since spatial inversion is an isometry (and thus bi-Lipschitz), it strictly preserves dimension, yielding $\dim_H(-\mathcal{M}) = \dim_H(\mathcal{M}) \leq k$.

The chord set $\mathcal{C}$ corresponds to the Minkowski difference, equivalent to the set addition $\mathcal{M} + (-\mathcal{M})$. This operation maps the Cartesian product $\mathcal{M} \times (-\mathcal{M})$ under the Lipschitz continuous addition function $f$: 
\begin{equation} 
f: \mathcal{M} \times (-\mathcal{M}) \rightarrow \mathbb{R}^n, \quad f(\mathbf{x}, \mathbf{y}) = \mathbf{x} + \mathbf{y}. 
\end{equation} 
Applying the standard dimension inequality for Lipschitz mappings and Cartesian products \cite[Formula 7.5]{falconer2014fractal}, we obtain: 
\begin{align} 
\dim_H(\mathcal{C}) &= \dim_H(f(\mathcal{M} \times (-\mathcal{M}))) \leq \dim_H(\mathcal{M} \times (-\mathcal{M})) \nonumber \\ 
&\leq \dim_H(\mathcal{M}) + \dim_H(-\mathcal{M}). \label{eq:dim_bound} 
\end{align} 
Substituting the initial bounds into \eqref{eq:dim_bound} yields $\dim_H(\mathcal{C}) \leq k + k = 2k$. \hfill $\blacksquare$

\vspace{1.5mm} 
\noindent\textbf{Adversarial Synthesis.} 
Drawing upon residual learning principles \cite{he2016deep}, the final adversarial sample $\mathbf{x}_{\mathrm{final}}$ is synthesized by translating the pristine input $\mathbf{x}_0$ by the chord vector $\Delta$: 
\begin{equation} 
\mathbf{x}_{\mathrm{final}} = \mathbf{x}_0 + \Delta. 
\end{equation} 
By applying this additive residual, the search space shifts from the heavily constrained $k$-dimensional manifold $\mathcal{M}$ to a richer $2k$-dimensional space $\mathbf{x}_0 + \mathcal{C}$ centered at $\mathbf{x}_0$. This enables the direct injection of semantic features, largely mitigating the inherent reconstruction constraints of the autoencoder.

Crucially, the RAG mechanism independently addresses two core limitations of standard latent optimization. First, it anchors the semantic shift directly to the pristine input $\mathbf{x}_0$ via residual addition, mitigating decoding artifacts to guarantee visual fidelity. Second, it formulates the perturbation as a geometric chord, expanding the search space up to $2k$ dimensions to accelerate boundary navigation and maximize query efficiency, while preserving the semantic meaningfulness inherently tied to the original manifold.

\begin{algorithm}
\caption{LGC (Latent Geometric Chords)}
\label{alg:lgc}
\begin{algorithmic}[1]
\STATE \textbf{Inputs:} Source image $\mathbf{x}_0$, indicator function $\phi(\cdot)$, Autoencoder $(E(\cdot), G(\cdot))$, random direction $\Theta$, queries to estimate initial normal vector $N_0$, iteration $T$.
\STATE \textbf{Output:} Adversarial example $\mathbf{x}_{\mathrm{final}}$.
\STATE $\mathbf{z}_s \leftarrow E(\mathbf{x}_0)$
\STATE Define residual mapping $\Phi(\mathbf{z}) = \text{clip}_{[0,1]^n}\left(\mathbf{x}_0 + G(\mathbf{z}) - G(\mathbf{z}_s)\right)$
\STATE $r \leftarrow \min\left\{r > 0 : \phi\left(\Phi\left(\mathbf{z}_s + r * \frac{\Theta}{\|\Theta\|_2}\right)\right) = 1\right\}$
\STATE $\mathbf{z}_{b_1} \leftarrow \text{BinarySearch}\left(\mathbf{z}_s, \mathbf{z}_s + r * \frac{\Theta}{\|\Theta\|_2}, \phi \circ \Phi\right)$ \COMMENT{Find initial boundary point}
\FOR{$t=1$ \textbf{to} $T$}
    \STATE Generate $N_t = N_0 \sqrt{t}$ samples, $\mathbf{u}_k \sim \mathcal{N}(0, \sigma^2\mathbf{I})$
    \STATE Estimate $\hat{\mathbf{n}}_t$ using $\mathbf{u}_k$ at $\mathbf{z}_{b_t}$ by $N_t$ queries via $\Phi(\cdot)$.
    \STATE $\hat{\mathbf{v}}_t = \frac{\mathbf{z}_{b_t} - \mathbf{z}_s}{\|\mathbf{z}_{b_t} - \mathbf{z}_s\|_2}$
    \STATE $\theta_t = \cos^{-1}(\hat{\mathbf{n}}_t \cdot \hat{\mathbf{v}}_t)$, \quad $i = 1$
    \WHILE{True}
        \STATE $m_i = \sin \theta_t \tan\left(\frac{90^\circ}{2^i}\right) - \cos \theta_t$
        \STATE $\hat{\zeta}_t = (\hat{\mathbf{n}}_t + m_i \hat{\mathbf{v}}_t)/\|\hat{\mathbf{n}}_t + m_i \hat{\mathbf{v}}_t\|_2$
        \STATE $\mathbf{z}_q = \mathbf{z}_s + \|\mathbf{z}_{b_t} - \mathbf{z}_s\|_2(\hat{\zeta}_t \cdot \hat{\mathbf{v}}_t)\hat{\zeta}_t$, \quad $i = i + 1$
        \IF{$\phi(\Phi(\mathbf{z}_q)) = -1$}
            \STATE \textbf{break}
        \ENDIF
    \ENDWHILE
    \STATE $\mathbf{z}_{b_{t+1}} \leftarrow \text{BSSP}(\mathbf{z}_s, \mathbf{z}_q, \mathbf{z}_{b_t}, \phi \circ \Phi)$ \COMMENT{Find boundary point on semicircular path}
\ENDFOR
\STATE $\mathbf{x}_{\mathrm{final}} \leftarrow \Phi(\mathbf{z}_{b_{T+1}})$
\end{algorithmic}
\end{algorithm}

\begin{figure}[t]
    \centering
    \includegraphics[width=0.5\textwidth]{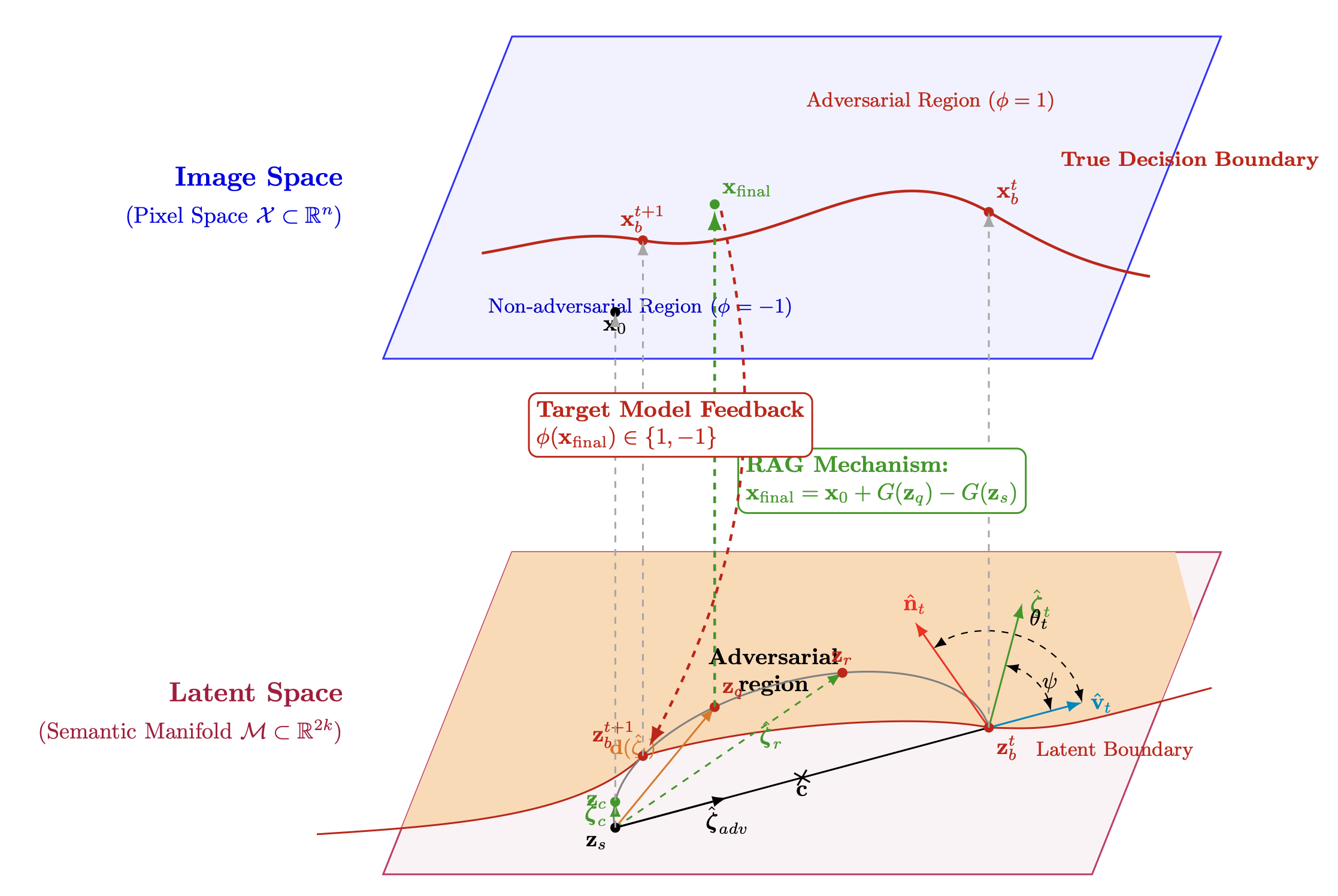}
    \caption{Dual-space architecture of the Latent Geometric Chords (LGC) method. The geometric boundary search ($\hat{\zeta}_c, \hat{\zeta}_r$) and continuous bisection are strictly confined to the 2D plane within the semantic manifold (bottom). The exact decision boundary crossing is evaluated in the image space (top) using the Residual-based Adversarial Generation (RAG) mechanism.}
    \label{fig:lgc_architecture}
\end{figure}

\subsection{LGC}
The entire step-by-step procedure of our proposed LGC is outlined in Algorithm \ref{alg:lgc}. Let $\mathbf{z}_s$ serve as the origin within the semantic manifold, while $\mathbf{z}_b^t$ denotes the established boundary intersection at step $t$. We define a normalized directional vector $\hat{\mathbf{v}}_t$ pointing from the baseline $\mathbf{z}_s$ to boundary, computed as $\hat{\mathbf{v}}_t = (\mathbf{z}_b^t - \mathbf{z}_s) / \|\mathbf{z}_b^t - \mathbf{z}_s\|_2$. Letting $\hat{\mathbf{n}}_t$ represent the estimated outward normal at $\mathbf{z}_b^t$, our strategy executes a tightly constrained search to locate optimal subsequent boundary point $\mathbf{z}_b^{t+1}$ in the semantic manifold. This local exploration is dynamically confined to a semicircular arc residing entirely within the 2D geometric plane spanned by $\{\hat{\mathbf{n}}_t, \hat{\mathbf{v}}_t\}$. To ensure directional consistency, the search is restricted to the subspace proximate to $\hat{\mathbf{n}}_t$. The trajectory is geometrically defined by its endpoints, $\mathbf{z}_s$ and $\mathbf{z}_b^t$, and is centered at the midpoint:
 \begin{equation}
\mathbf{c} = \frac{\mathbf{z}_b^t + \mathbf{z}_s}{2}.
 \end{equation}

The path maintains a constant radius $R$, defined by the Euclidean distance:
\begin{equation}
R = \frac{\|\mathbf{z}_b^t - \mathbf{z}_s\|_2}{2}.
\end{equation}

The search direction $\hat{\zeta}_t(m)$ within this plane is parameterized by a scalar factor $m$:
\begin{equation}
\hat{\zeta}_t(m) = \frac{\hat{\mathbf{n}}_t + m\hat{\mathbf{v}}_t}{\|\hat{\mathbf{n}}_t + m\hat{\mathbf{v}}_t\|_2}.
\end{equation}

For a given $\hat{\zeta}_t$, the corresponding perturbation vector $\mathbf{d}(\hat{\zeta}_t)$ that projects the query onto the semicircular arc is:
\begin{equation}
\mathbf{d}(\hat{\zeta}_t) = \|\mathbf{z}_b^t - \mathbf{z}_s\|_2 (\hat{\zeta}_t \cdot \hat{\mathbf{v}}_t) \hat{\zeta}_t = \|\mathbf{z}_b^t - \mathbf{z}_s\|_2 \cos \psi \cdot \hat{\zeta}_t.
\end{equation}
where $\psi$ is the angular displacement between $\hat{\zeta}_t$ and $\hat{\mathbf{v}}_t$. Then, the latent adversarial vector $\mathbf{z}_q$ is generated using the formula $\mathbf{z}_q = \mathbf{z}_s + \mathbf{d}(\hat{\zeta}_t)$. This vector $\mathbf{z}_q$ is reconstructed into the image space to synthesize the final adversarial image, $\mathbf{x}_{\mathrm{final}}$, using the Residual-based Adversarial Generation (RAG) method. The final adversarial image is evaluated by the targeted black-box classification model in the image space. Similar to the CGBA method, we query the target model with $\mathbf{x}_{\mathrm{final}}$ to guide the boundary search along the semicircular path within the latent space. 

The boundary search is seamlessly executed along this semicircular trajectory within the latent domain. Specifically, the process initially identifies two latent vectors—one non-adversarial and one adversarial—residing on the semicircular path. While their adversarial status is evaluated by generating adversarial images via the RAG method in the pixel space, the search process itself is conducted strictly within the latent manifold. Inspired by binary search techniques in the CGBA method, the algorithm iteratively narrows the latent interval between these two vectors. This progressive refinement converges upon the subsequent latent boundary point $\mathbf{z}_b^{t+1}$, which mathematically corresponds to the pixel-space boundary point $\mathbf{x}_b^{t+1}$.

By utilizing the trigonometric identity $\cot(90^\circ - \alpha) = \tan(\alpha)$, the multiplication factor $m_i$ required to attain the progressively refined search angle is simplified as:
\begin{equation}
m_i = \sin \theta_t \tan\left(\frac{90^\circ}{2^i}\right) - \cos \theta_t, \quad \forall i \in \mathbb{Z}^+.
\end{equation}
where $\theta_t = \cos^{-1} (\hat{\mathbf{v}}_t \cdot \hat{\mathbf{n}}_t)$. As $i$ increases, the search angle monotonically increases. By iterating through specific values of $m_i$, a perturbation vector $\mathbf{d}(\hat{\zeta}_t(m_i))$ is computed to identify a corresponding non-adversarial candidate point $\mathbf{x}_c$ such that $\phi(\mathbf{x}_c) = -1$. This image-space point is defined by the transformation:
\begin{equation}
\mathbf{x}_c = \mathbf{x}_0 + \left(G(\mathbf{z}_s + \mathbf{d}(\hat{\zeta}_t(m_i))) - G(\mathbf{z}_s)\right).
\end{equation}

Subsequently, a boundary search is initiated between the non-adversarial latent point $\mathbf{z}_c$ and the current adversarial candidate $\mathbf{z}_b^t$ , where $\mathbf{z}_c = \mathbf{z}_s + \mathbf{d}(\hat{\zeta}_t(m_i))$. This search utilizes the \textit{Boundary Search along a Semicircular Path} (BSSP) algorithm to isolate $\mathbf{z}_b^{t+1}$ along the latent trajectory. Specifically, BSSP operates by evaluating the midpoint of the angular interval between a known adversarial and non-adversarial latent vector. Based on the classifier's hard-label feedback at this midpoint, it iteratively halves the search interval, effectively converging to the precise decision boundary point with minimal queries. At the commencement of the BSSP phase for step $t$, let $\hat{\zeta}_{\mathrm{adv}}$ and $\hat{\zeta}_c$ denote the unit directions of $\mathbf{z}_b^t$ and $\mathbf{z}_c$ relative to $\mathbf{z}_s$. The resultant bisecting search direction, $\hat{\zeta}_r$, is defined as the normalized vector sum:
\begin{equation}
\hat{\zeta}_r = \frac{\hat{\zeta}_{\mathrm{adv}} + \hat{\zeta}_c}{\|\hat{\zeta}_{\mathrm{adv}} + \hat{\zeta}_c\|_2}.
\end{equation}

The BSSP iteratively narrows the angular search space. If the decision function yields $\phi(\mathbf{x}_r) = 1$ (adversarial), the search interval $[\hat{\zeta}_{\mathrm{adv}}, \hat{\zeta}_c]$ is bisected and reduced to $[\hat{\zeta}_r, \hat{\zeta}_c]$, as the optimal boundary point $\mathbf{z}_b^{t+1}$ must lie within this range. Conversely, if $\phi(\mathbf{x}_r) = -1$ (non-adversarial), the interval is refined to $[\hat{\zeta}_{\mathrm{adv}}, \hat{\zeta}_r]$, where:
\begin{equation}
\mathbf{x}_r = \mathbf{x}_0 + \left(G(\mathbf{z}_s + \mathbf{d}(\hat{\zeta}_r)) - G(\mathbf{z}_s)\right).
\end{equation}

This bisection process continues until $\mathbf{z}_b^{t+1}$ is located with a predefined level of precision. A primary mathematical advantage of the BSSP is its ability to guarantee a continuous reduction in perturbation magnitude; specifically, for any query point $\mathbf{z}_q$ located on the semicircular path, the geometric condition $\|\mathbf{z}_q - \mathbf{z}_s\|_2 \leq \|\mathbf{z}_b^t - \mathbf{z}_s\|_2$ is inherently maintained.

\begin{algorithm}
\caption{LGC-H}
\label{alg:lgc_h}
\begin{algorithmic}[1]
\STATE \textbf{Inputs:} Source image $\mathbf{x}_0$, target image/random noise $\mathbf{x}_t$ (to find initial adversarial class), indicator function $\phi(\cdot)$, Autoencoder $(E(\cdot), G(\cdot))$, queries to find initial normal vector $N_0$, iteration $T$.
\STATE \textbf{Output:} Adversarial example $\mathbf{x}_{\mathrm{final}}$.
\STATE $\mathbf{z}_s \leftarrow E(\mathbf{x}_0)$
\STATE Define residual mapping $\Phi(\mathbf{z}) = \text{clip}_{[0,1]^n}\left(\mathbf{x}_0 + G(\mathbf{z}) - G(\mathbf{z}_s)\right)$
\STATE Find initial adversarial latent vector $\mathbf{z}_{\mathrm{adv}}$
\STATE $\mathbf{z}_{b_1} \leftarrow \text{BinarySearch}(\mathbf{z}_s, \mathbf{z}_{\mathrm{adv}}, \phi \circ \Phi)$ \COMMENT{Find initial boundary point}
\FOR{$t=1$ \textbf{to} $T$}
    \STATE Generate $N_t = N_0 \sqrt{t}$ samples, $\mathbf{u}_k \sim \mathcal{N}(0, \sigma^2\mathbf{I})$
    \STATE Estimate $\hat{\mathbf{n}}_t$ using $\mathbf{u}_k$ at $\mathbf{z}_{b_t}$ by $N_t$ queries via $\Phi(\cdot)$.
    \STATE $\hat{\mathbf{v}}_t = \frac{\mathbf{z}_{b_t} - \mathbf{z}_s}{\|\mathbf{z}_{b_t} - \mathbf{z}_s\|_2}$
    \STATE $\theta_t = \cos^{-1}(\hat{\mathbf{n}}_t \cdot \hat{\mathbf{v}}_t)$, \quad $i = 1$
    \WHILE{True}
        \STATE $m_i = \sin\theta_t \cot\left(\frac{\theta_t}{2^i}\right) - \cos\theta_t$
        \STATE $\hat{\zeta}_t = (\hat{\mathbf{n}}_t + m_i \hat{\mathbf{v}}_t)/\|\hat{\mathbf{n}}_t + m_i \hat{\mathbf{v}}_t\|_2$
        \STATE $\mathbf{z}_h = \mathbf{z}_s + \|\mathbf{z}_{b_t} - \mathbf{z}_s\|_2(\hat{\zeta}_t \cdot \hat{\mathbf{v}}_t)\hat{\zeta}_t$, \quad $i = i + 1$
        \IF{$\phi(\Phi(\mathbf{z}_h)) = 1$}
            \STATE \textbf{break}
        \ENDIF
    \ENDWHILE
    \STATE $\mathbf{z}_{b_{t+1}} \leftarrow \text{BinarySearch}(\mathbf{z}_s, \mathbf{z}_h, \phi \circ \Phi)$ \COMMENT{Find boundary point}
\ENDFOR
\STATE $\mathbf{x}_{\mathrm{final}} \leftarrow \Phi(\mathbf{z}_{b_{T+1}})$
\end{algorithmic}
\end{algorithm}

\subsection{LGC-H}

While standard geometric searches are effective for low-curvature topologies, targeted attacks frequently produce highly curved boundaries with narrow adversarial regions. To overcome the severe query inefficiency of conventional semicircular sweeps in these scenarios, we propose LGC-H, an accelerated variant designed to rapidly isolate adversarial trajectories within the compressed semantic manifold. 

The primary contribution of LGC-H is integrating a dynamic angular bisection strategy directly into the latent domain. Although trigonometric bisection has been explored for image-space perturbations \cite{reza2023cgba}, applying it to a highly non-linear latent topology typically induces severe reconstruction errors. We resolve this by strictly coupling geometrically bisected latent paths with our Residual-based Adversarial Generation (RAG) mechanism to effectively neutralize decoder distortions.

Let $\theta_t = \arccos(\hat{\mathbf{n}}_t^\top \hat{\mathbf{v}}_t)$ denote the angle between the estimated latent gradient $\hat{\mathbf{n}}_t$ and the baseline projection $\hat{\mathbf{v}}_t$. To dynamically adjust the search resolution without exhaustive fixed-step evaluations, we introduce a scaling coefficient $m_i$:
\begin{equation}
m_i = \sin\theta_t \cot\left(\frac{\theta_t}{2^i}\right) - \cos\theta_t, \quad \forall i \in \mathbb{Z}^+.
\end{equation}
This formulation ensures the updated perturbation trajectory $\hat{\zeta}_t(m_i)$ maintains a precise angular offset of $\theta_t / 2^i$ relative to $\hat{\mathbf{v}}_t$. Incrementing the discrete step parameter $i$ monotonically halves the search space, enabling exponential convergence on viable adversarial subspaces.

Once an optimal coefficient $m_i$ identifies an adversarial latent coordinate $\mathbf{z}_h = \mathbf{z}_s + \mathbf{d}(\hat{\zeta}_t(m_i))$ (where $\phi(\mathbf{x}_h) = 1$), our RAG module directly overlays the perturbation chord onto the source image to strictly preserve semantic aesthetics:
\begin{equation}
\mathbf{x}_h = \mathbf{x}_0 + \left(G(\mathbf{z}_h) - G(\mathbf{z}_s)\right).
\end{equation}
Finally, a localized binary search between the non-adversarial anchor $\mathbf{z}_s$ and the adversarial candidate $\mathbf{z}_h$ refines the exact boundary transition $\mathbf{z}_b^{t+1}$. By confining geometric bisection to the latent manifold and leveraging RAG for reconstruction, LGC-H achieves outstanding query efficiency without sacrificing structural integrity.

\subsection{Initialization for Targeted Latent-Space Attacks}

Directly encoding a target image for initialization is ineffective because adding the source image's residual during RAG reconstruction distorts the intended semantics. To overcome this, we introduce a ``pseudo-target'' strategy. 

We first isolate the pixel-space residual $\mathbf{R} = \mathbf{x}_0 - G(\mathbf{z}_s)$. To effectively cancel out reconstruction artifacts, we subtract this residual from the target image $\mathbf{x}_{\mathrm{target}}$ to synthesize a pseudo-target, which is then mapped to the latent space:
\begin{equation}
\mathbf{z}_{\mathrm{target}} = E\left(\text{clip}_{[0,1]^n}(\mathbf{x}_{\mathrm{target}} - \mathbf{R})\right).
\end{equation}

Due to the highly non-linear generative manifold, this vector may occasionally deviate from the intended target class. If this occurs, we briefly fine-tune $\mathbf{z}_{\mathrm{target}}$ via the Adam optimizer (e.g., learning rate of 0.05, up to 50 iterations) to minimize the cross-entropy loss and guarantee adversarial validity. Finally, a standard binary search between $\mathbf{z}_s$ and $\mathbf{z}_{\mathrm{target}}$ locates the precise initial boundary point $\mathbf{z}_b^1$ to commence the semicircular boundary search.


\begin{table*}[t]
\centering
\scriptsize
\setlength{\tabcolsep}{5.5pt} 
\renewcommand{\arraystretch}{0.8} 
\setlength{\aboverulesep}{0.5pt} 
\setlength{\belowrulesep}{0.5pt} 
\vspace{-5mm}

\caption{Average SSIM / LPIPS values of perturbation for targeted and non-targeted black-box attacks under different query budgets.}
\vspace{-2mm}
\label{tab:ssim_lpips_results}
\resizebox{\textwidth}{!}{
\begin{tabular}{lll cccc cccc}
\toprule
 & & & \multicolumn{4}{c}{Non-targeted} & \multicolumn{4}{c}{Targeted} \\[-0.01ex]
\cmidrule(lr){4-7} \cmidrule(lr){8-11}
Dataset & Model & Attack & 1K & 2.5K & 5K & 10K & 1K & 2.5K & 5K & 10K \\
\midrule
\multirow[c]{12}{*}{\rotatebox[origin=c]{90}{\textbf{Places365}}} & \multirow[c]{6}{*}{ResNet50} & Sign\_OPT & 0.645 / 0.311 & 0.735 / 0.244 & 0.799 / 0.197 & 0.852 / 0.157 & 0.390 / 0.582 & 0.426 / 0.548 & 0.470 / 0.507 & 0.516 / 0.464 \\
 & & HSJA & 0.764 / 0.212 & 0.852 / 0.147 & 0.901 / 0.109 & 0.939 / 0.077 & 0.425 / 0.536 & 0.487 / 0.475 & 0.561 / 0.405 & 0.680 / 0.303 \\
 & & CGBA-H & 0.757 / 0.270 & 0.809 / 0.226 & 0.853 / 0.187 & 0.902 / 0.138 & 0.481 / 0.534 & 0.508 / 0.519 & 0.540 / 0.499 & 0.589 / 0.464 \\
 & & CGBA & 0.785 / 0.261 & 0.836 / 0.215 & 0.880 / 0.170 & 0.919 / 0.124 & 0.388 / 0.608 & 0.367 / 0.627 & 0.350 / 0.641 & 0.337 / 0.651 \\
 & & LGC-H & 0.978 / 0.043 & 0.992 / 0.019 & \textbf{0.996} / 0.011 & 0.997 / \textbf{0.008} & 0.723 / 0.326 & 0.826 / 0.225 & 0.897 / 0.149 & 0.945 / 0.095 \\
 & & LGC & \textbf{0.980} / \textbf{0.039} & \textbf{0.993} / \textbf{0.016} & \textbf{0.996} / \textbf{0.010} & \textbf{0.998} / \textbf{0.008} & \textbf{0.740} / \textbf{0.319} & \textbf{0.865} / \textbf{0.194} & \textbf{0.933} / \textbf{0.116} & \textbf{0.970} / \textbf{0.069} \\
\cmidrule{2-11}
 & \multirow[c]{6}{*}{DenseNet161} & Sign\_OPT & 0.641 / 0.309 & 0.742 / 0.235 & 0.810 / 0.186 & 0.867 / 0.143 & 0.389 / 0.575 & 0.421 / 0.545 & 0.458 / 0.510 & 0.501 / 0.469 \\
 & & HSJA & 0.755 / 0.219 & 0.847 / 0.151 & 0.898 / 0.112 & 0.937 / 0.079 & 0.425 / 0.535 & 0.486 / 0.478 & 0.563 / 0.406 & 0.684 / 0.300 \\
 & & CGBA-H & 0.752 / 0.273 & 0.811 / 0.224 & 0.865 / 0.173 & 0.911 / 0.124 & 0.489 / 0.537 & 0.522 / 0.518 & 0.561 / 0.492 & 0.623 / 0.446 \\
 & & CGBA & 0.786 / 0.262 & 0.839 / 0.209 & 0.889 / 0.158 & 0.932 / 0.110 & 0.383 / 0.615 & 0.364 / 0.631 & 0.353 / 0.639 & 0.344 / 0.643 \\
 & & LGC-H & 0.976 / 0.046 & 0.991 / 0.021 & \textbf{0.996} / 0.012 & \textbf{0.997} / \textbf{0.008} & 0.726 / 0.317 & 0.832 / 0.215 & 0.907 / 0.135 & 0.951 / 0.083 \\
 & & LGC & \textbf{0.979} / \textbf{0.042} & \textbf{0.993} / \textbf{0.018} & \textbf{0.996} / \textbf{0.010} & \textbf{0.997} / \textbf{0.008} & \textbf{0.739} / \textbf{0.312} & \textbf{0.869} / \textbf{0.187} & \textbf{0.938} / \textbf{0.108} & \textbf{0.973} / \textbf{0.062} \\
\midrule
\multirow[c]{24}{*}{\rotatebox[origin=c]{90}{\textbf{ImageNet}}} & \multirow[c]{6}{*}{ResNet50} & Sign\_OPT & 0.463 / 0.428 & 0.683 / 0.261 & 0.846 / 0.143 & 0.940 / 0.069 & 0.417 / 0.591 & 0.449 / 0.562 & 0.483 / 0.530 & 0.520 / 0.493 \\
 & & HSJA & 0.546 / 0.387 & 0.739 / 0.241 & 0.860 / 0.147 & 0.936 / 0.082 & 0.449 / 0.547 & 0.491 / 0.499 & 0.569 / 0.424 & 0.726 / 0.290 \\
 & & CGBA-H & 0.658 / 0.343 & 0.735 / 0.284 & 0.810 / 0.221 & 0.874 / 0.158 & 0.506 / 0.525 & 0.539 / 0.501 & 0.581 / 0.468 & 0.634 / 0.422 \\
 & & CGBA & 0.682 / 0.342 & 0.757 / 0.279 & 0.824 / 0.217 & 0.892 / 0.147 & 0.373 / 0.624 & 0.345 / 0.643 & 0.330 / 0.652 & 0.326 / 0.651 \\
 & & LGC-H & 0.947 / 0.093 & 0.980 / 0.042 & 0.990 / 0.023 & 0.995 / \textbf{0.014} & 0.698 / \textbf{0.355} & 0.792 / 0.261 & 0.870 / 0.176 & 0.930 / 0.105 \\
 & & LGC & \textbf{0.977} / \textbf{0.049} & \textbf{0.993} / \textbf{0.023} & \textbf{0.996} / \textbf{0.017} & \textbf{0.997} / 0.015 & \textbf{0.701} / 0.358 & \textbf{0.819} / \textbf{0.242} & \textbf{0.906} / \textbf{0.144} & \textbf{0.960} / \textbf{0.075} \\
\cmidrule{2-11}
 & \multirow[c]{6}{*}{DenseNet121} & Sign\_OPT & 0.551 / 0.360 & 0.786 / 0.188 & 0.914 / 0.091 & 0.972 / 0.040 & 0.411 / 0.581 & 0.457 / 0.535 & 0.502 / 0.492 & 0.554 / 0.443 \\
 & & HSJA & 0.674 / 0.287 & 0.854 / 0.153 & 0.928 / 0.092 & 0.966 / 0.054 & 0.404 / 0.563 & 0.447 / 0.506 & 0.565 / 0.403 & 0.743 / 0.261 \\
 & & CGBA-H & 0.678 / 0.334 & 0.763 / 0.262 & 0.831 / 0.200 & 0.893 / 0.137 & 0.509 / 0.539 & 0.542 / 0.513 & 0.579 / 0.483 & 0.634 / 0.435 \\
 & & CGBA & 0.694 / 0.336 & 0.780 / 0.261 & 0.845 / 0.195 & 0.909 / 0.126 & 0.362 / 0.644 & 0.333 / 0.658 & 0.320 / 0.662 & 0.321 / 0.655 \\
 & & LGC-H & \textbf{0.971} / 0.055 & 0.991 / 0.021 & \textbf{0.996} / 0.012 & 0.997 / 0.008 & 0.730 / \textbf{0.323} & 0.830 / 0.217 & 0.906 / 0.131 & 0.955 / 0.074 \\
 & & LGC & \textbf{0.971} / \textbf{0.054} & \textbf{0.992} / \textbf{0.019} & \textbf{0.996} / \textbf{0.010} & \textbf{0.998} / \textbf{0.007} & \textbf{0.732} / 0.327 & \textbf{0.863} / \textbf{0.191} & \textbf{0.939} / \textbf{0.102} & \textbf{0.977} / \textbf{0.051} \\
\cmidrule{2-11}
 & \multirow[c]{6}{*}{VGG16} & Sign\_OPT & 0.712 / 0.248 & 0.873 / 0.130 & 0.951 / 0.068 & 0.983 / 0.036 & 0.424 / 0.585 & 0.499 / 0.522 & 0.567 / 0.464 & 0.628 / 0.410 \\
 & & HSJA & 0.801 / 0.198 & 0.911 / 0.112 & 0.958 / 0.069 & 0.981 / 0.044 & 0.461 / 0.539 & 0.548 / 0.460 & 0.658 / 0.368 & 0.809 / 0.250 \\
 & & CGBA-H & 0.819 / 0.209 & 0.881 / 0.151 & 0.931 / 0.104 & 0.967 / 0.062 & 0.536 / 0.516 & 0.596 / 0.472 & 0.667 / 0.414 & 0.749 / 0.339 \\
 & & CGBA & 0.824 / 0.212 & 0.893 / 0.146 & 0.937 / 0.099 & 0.970 / 0.059 & 0.326 / 0.659 & 0.302 / 0.672 & 0.297 / 0.673 & 0.307 / 0.661 \\
 & & LGC-H & \textbf{0.977} / \textbf{0.051} & \textbf{0.991} / \textbf{0.027} & \textbf{0.995} / 0.020 & \textbf{0.997} / 0.017 & 0.713 / 0.351 & 0.818 / 0.252 & 0.895 / 0.173 & 0.946 / 0.118 \\
 & & LGC & 0.949 / 0.089 & 0.981 / 0.038 & 0.992 / \textbf{0.019} & 0.996 / \textbf{0.012} & \textbf{0.719} / \textbf{0.348} & \textbf{0.851} / \textbf{0.224} & \textbf{0.932} / \textbf{0.140} & \textbf{0.972} / \textbf{0.091} \\
\cmidrule{2-11}
 & \multirow[c]{6}{*}{ViT} & Sign\_OPT & 0.428 / 0.459 & 0.722 / 0.234 & 0.886 / 0.118 & 0.957 / 0.057 & 0.430 / 0.559 & 0.468 / 0.516 & 0.505 / 0.480 & 0.545 / 0.443 \\
 & & HSJA & 0.556 / 0.382 & 0.778 / 0.214 & 0.883 / 0.132 & 0.940 / 0.081 & 0.429 / 0.513 & 0.520 / 0.424 & 0.684 / 0.297 & 0.847 / 0.172 \\
 & & CGBA-H & 0.825 / 0.219 & 0.944 / 0.092 & 0.979 / 0.044 & 0.992 / 0.022 & 0.620 / 0.445 & 0.714 / 0.357 & 0.812 / 0.258 & 0.896 / 0.158 \\
 & & CGBA & 0.827 / 0.225 & 0.946 / 0.094 & 0.981 / 0.043 & 0.993 / 0.019 & 0.412 / 0.597 & 0.515 / 0.513 & 0.677 / 0.374 & 0.841 / 0.210 \\
 & & LGC-H & 0.955 / 0.081 & 0.986 / 0.031 & 0.993 / 0.017 & \textbf{0.996} / 0.012 & 0.804 / 0.241 & 0.888 / 0.146 & 0.942 / 0.081 & 0.973 / 0.043 \\
 & & LGC & \textbf{0.956} / \textbf{0.079} & \textbf{0.988} / \textbf{0.027} & \textbf{0.994} / \textbf{0.015} & \textbf{0.996} / \textbf{0.010} & \textbf{0.813} / \textbf{0.236} & \textbf{0.916} / \textbf{0.122} & \textbf{0.967} / \textbf{0.058} & \textbf{0.988} / \textbf{0.029} \\
\bottomrule
\end{tabular}
}
\end{table*}

\begin{table*}[t]
\centering
\scriptsize
\setlength{\tabcolsep}{5.5pt} 
\renewcommand{\arraystretch}{0.8} 
\setlength{\aboverulesep}{0.5pt} 
\setlength{\belowrulesep}{0.5pt}

\caption{Average (median) $L_{2}$ norm of perturbation for targeted and non-targeted black-box attacks under different query budgets}
\label{tab:main_results}
\resizebox{\textwidth}{!}{
\begin{tabular}{lll cccc cccc}
\toprule
 & & & \multicolumn{4}{c}{Non-targeted attacks} & \multicolumn{4}{c}{Targeted attacks} \\[-0.001ex]
\cmidrule(lr){4-7} \cmidrule(lr){8-11}
Dataset & Model & Attack & 1000 QUE & 2500 QUE & 5000 QUE & 10000 QUE & 1000 QUE & 2500 QUE & 5000 QUE & 10000 QUE \\
\midrule
\multirow[c]{12}{*}{\rotatebox[origin=c]{90}{\textbf{Places365}}} & \multirow[c]{6}{*}{ResNet50} & Sign\_OPT & 22.138(19.489) & 15.983(14.442) & 12.349(11.452) & 9.588(8.993) & 89.543(85.536) & 78.607(70.366) & 69.446(57.076) & 60.841(44.985) \\
 & & HSJA & 16.331(14.057) & 10.907(9.385) & 7.983(7.089) & 5.799(5.365) & 71.591(68.533) & 56.261(52.072) & 43.233(38.528) & 28.351(22.192) \\
 & & CGBA-H & 16.539(13.103) & 13.489(10.178) & 10.909(8.001) & 7.999(5.356) & 79.347(78.646) & 73.404(73.612) & 67.186(66.401) & 58.135(59.783) \\
 & & CGBA & 15.338(13.159) & 12.343(9.548) & 9.704(6.804) & 7.163(4.578) & 88.283(86.054) & 87.222(85.334) & 85.737(83.937) & 83.097(81.930) \\
 & & LGC-H & 7.393(5.542) & 4.351(3.286) & 3.069(2.382) & 2.465(1.996) & 60.885(58.091) & 41.365(37.726) & 27.362(23.576) & 17.088(14.349) \\
 & & LGC & \textbf{6.938(5.079)} & \textbf{3.906(2.914)} & \textbf{2.892(2.232)} & \textbf{2.401(1.955)} & \textbf{55.609(52.296)} & \textbf{31.574(28.043)} & \textbf{18.035(14.981)} & \textbf{10.185(8.977)} \\
\cmidrule{2-11}
 & \multirow[c]{6}{*}{DenseNet161} & Sign\_OPT & 22.706(20.610) & 15.833(14.031) & 11.885(10.975) & 8.883(8.014) & 90.605(80.851) & 80.487(65.588) & 72.732(53.973) & 65.214(43.221) \\
 & & HSJA & 16.737(14.128) & 11.174(9.292) & 8.244(6.877) & 5.916(5.127) & 70.288(69.774) & 56.189(55.854) & 43.114(40.523) & 27.829(23.722) \\
 & & CGBA-H & 17.395(13.675) & 13.883(11.046) & 10.614(6.827) & 7.745(4.171) & 75.677(73.333) & 68.738(66.557) & 61.031(58.175) & 50.480(46.865) \\
 & & CGBA & 15.627(11.713) & 12.408(8.698) & 9.300(5.912) & 6.589(3.913) & 87.511(86.447) & 85.916(85.533) & 83.880(84.073) & 80.862(81.957) \\
 & & LGC-H & 7.686(5.355) & 4.493(3.031) & 3.150(2.245) & 2.540(1.879) & 60.292(59.818) & 39.872(37.682) & 25.171(22.382) & 15.337(13.062) \\
 & & LGC & \textbf{7.184(5.133)} & \textbf{4.062(2.806)} & \textbf{2.907(2.080)} & \textbf{2.385(1.757)} & \textbf{55.192(54.785)} & \textbf{30.202(27.832)} & \textbf{16.807(14.112)} & \textbf{9.497(8.164)} \\
\midrule
\multirow[c]{24}{*}{\rotatebox[origin=c]{90}{\textbf{ImageNet}}} & \multirow[c]{6}{*}{ResNet50} & Sign\_OPT & 40.032(34.904) & 20.028(15.578) & 10.037(6.851) & 4.721(3.102) & 87.232(75.958) & 80.288(62.784) & 74.786(51.486) & 69.563(41.110) \\
 & & HSJA & 28.999(23.870) & 15.359(11.015) & 8.577(5.843) & 4.664(3.405) & \textbf{61.895(60.637)} & 49.585(48.052) & 37.004(34.495) & 21.634(18.181) \\
 & & CGBA-H & 24.009(20.793) & 18.743(15.124) & 14.209(10.147) & 10.205(5.790) & 76.479(74.281) & 68.805(67.578) & 60.142(60.410) & 50.432(50.854) \\
 & & CGBA & 23.049(19.860) & 18.188(14.851) & 13.916(9.415) & 9.451(5.320) & 88.454(87.158) & 85.714(84.021) & 82.973(80.907) & 77.884(74.097) \\
 & & LGC-H & 12.246(9.414) & 6.807(5.052) & 4.578(3.528) & 3.435(2.728) & 68.444(66.757) & 49.950(49.326) & 34.482(34.014) & 21.561(19.848) \\
 & & LGC & \textbf{7.311(4.541)} & \textbf{3.858(2.525)} & \textbf{2.846(1.986)} & \textbf{2.383(1.766)} & 65.933(64.207) & \textbf{41.853(40.316)} & \textbf{24.295(22.194)} & \textbf{12.846(11.161)} \\
\cmidrule{2-11}
 & \multirow[c]{6}{*}{DenseNet121} & Sign\_OPT & 31.931(26.644) & 14.182(10.491) & 6.626(4.590) & 3.245(2.255) & 85.794(71.605) & 74.501(54.758) & 66.925(42.276) & 59.846(32.835) \\
 & & HSJA & 20.066(16.071) & 9.515(7.056) & 5.777(4.428) & 3.709(3.125) & 66.407(63.307) & 50.479(48.074) & 34.418(30.039) & 18.535(15.862) \\
 & & CGBA-H & 23.582(18.987) & 17.538(13.616) & 13.014(8.936) & 9.070(4.831) & 75.011(70.757) & 67.521(62.749) & 60.003(56.107) & 50.200(47.120) \\
 & & CGBA & 22.829(18.981) & 16.926(12.677) & 12.604(7.776) & 8.275(4.190) & 89.062(86.657) & 85.871(80.925) & 82.749(78.499) & 77.448(75.387) \\
 & & LGC-H & 8.568(6.640) & 4.606(3.358) & 3.169(2.375) & 2.481(1.903) & 61.438(60.259) & 42.037(41.034) & 26.504(24.318) & 15.229(13.580) \\
 & & LGC & \textbf{8.409(6.015)} & \textbf{4.269(2.989)} & \textbf{2.923(2.239)} & \textbf{2.322(1.813)} & \textbf{58.742(57.671)} & \textbf{32.755(30.324)} & \textbf{17.333(15.143)} & \textbf{8.804(7.814)} \\
\cmidrule{2-11}
 & \multirow[c]{6}{*}{VGG16} & Sign\_OPT & 19.525(15.472) & 8.843(6.367) & 4.389(2.918) & \textbf{2.291(1.520)} & 89.230(80.593) & 74.703(55.650) & 63.931(38.204) & 55.778(24.796) \\
 & & HSJA & 12.723(7.856) & 6.448(4.095) & 3.938(2.765) & 2.550(1.912) & 65.946(62.845) & 46.969(44.628) & 31.434(28.874) & 17.303(12.547) \\
 & & CGBA-H & 13.573(8.762) & 9.665(5.122) & 6.474(2.795) & 3.934(1.417) & 69.810(69.328) & 57.296(55.771) & 45.569(43.213) & 33.085(28.736) \\
 & & CGBA & 13.492(9.023) & 8.987(4.694) & 5.986(2.616) & 3.571(1.407) & 93.138(89.213) & 90.510(86.318) & 86.896(84.035) & 81.261(77.516) \\
 & & LGC-H & \textbf{7.522(5.125)} & \textbf{4.301(2.846)} & \textbf{3.145(2.112)} & 2.577(1.832) & 65.728(63.740) & 44.508(42.078) & 28.558(26.499) & 17.281(15.628) \\
 & & LGC & 12.046(9.126) & 6.369(4.480) & 4.093(3.064) & 3.081(2.461) & \textbf{61.726(59.881)} & \textbf{34.545(32.673)} & \textbf{18.625(16.991)} & \textbf{9.995(8.786)} \\
\cmidrule{2-11}
 & \multirow[c]{6}{*}{ViT} & Sign\_OPT & 45.428(40.483) & 17.469(14.328) & 8.099(6.720) & 4.295(3.429) & 83.398(58.814) & 74.324(41.987) & 69.147(31.701) & 64.980(23.601) \\
 & & HSJA & 28.578(24.261) & 12.982(10.428) & 7.504(5.876) & 4.608(3.747) & 52.157(48.527) & 34.929(32.519) & 20.661(17.826) & 10.715(9.362) \\
 & & CGBA-H & 13.131(9.939) & 5.585(4.066) & 3.051(2.277) & 1.833(1.481) & 48.580(49.708) & 33.431(32.612) & 21.661(18.284) & 12.850(8.463) \\
 & & CGBA & 13.216(10.104) & \textbf{5.534(4.183)} & \textbf{2.961(2.244)} & \textbf{1.711(1.396)} & 60.515(61.621) & 42.015(40.581) & 26.575(21.461) & 13.714(8.287) \\
 & & LGC-H & 11.359(9.472) & 6.046(5.119) & 4.211(3.716) & 3.322(3.015) & 48.271(46.680) & 31.258(29.536) & 19.101(17.027) & 10.971(8.463) \\
 & & LGC & \textbf{11.041(8.850)} & 5.583(4.630) & 3.885(3.454) & 3.104(2.884) & \textbf{44.215(42.285)} & \textbf{23.328(22.017)} & \textbf{11.965(10.747)} & \textbf{6.337(5.631)} \\
\bottomrule
\end{tabular}
}
\end{table*}

\begin{table}[!t]
\centering
\tiny
\setlength{\tabcolsep}{5.5pt} 
\renewcommand{\arraystretch}{0.8} 
\setlength{\aboverulesep}{0.5pt} 
\setlength{\belowrulesep}{0.5pt}
  \caption{Average SIM and LPIPS of adversarial perturbations against ResNet-18 on CelebAMask-HQ for identity and gender classification across varying query budgets.}
\label{tab:combined_results}
\resizebox{\columnwidth}{!}{
\begin{tabular}{ll cccc}
\toprule
\multirow{2}{*}{\textbf{Method}} & \multirow{2}{*}{\textbf{Metric}} & \multicolumn{4}{c}{\textbf{Model Queries}} \\[-0.01ex]
\cmidrule(lr){3-6}
 & & 1000 & 3000 & 5000 & 10000 \\
\midrule
\multicolumn{4}{c}{\textbf{Identity Classification}} \\
\midrule
\multirow{2}{*}{HSJA \cite{chen2020hopskipjumpattack}} & SIM $\uparrow$ & 0.569 & 0.663 & 0.728 & 0.821 \\
 & LPIPS $\downarrow$ & 0.287 & 0.187 & 0.132 & 0.067 \\
\multirow{2}{*}{Latent-HSJA} & SIM $\uparrow$ & 0.719 & 0.779 & 0.797 & 0.811 \\
 & LPIPS $\downarrow$ & 0.119 & 0.072 & 0.061 & 0.053 \\
\multirow{2}{*}{LGC-H (Ours)} & SIM $\uparrow$ & 0.9837 & 0.9961 & 0.9984 & 0.9996 \\
 & LPIPS $\downarrow$ & 0.0818 & 0.0253 & 0.0123 & 0.0050 \\
\multirow{2}{*}{LGC (Ours)} & SIM $\uparrow$ & \textbf{0.9872} & \textbf{0.9985} & \textbf{0.9995} & \textbf{0.9999} \\
 & LPIPS $\downarrow$ & \textbf{0.0732} & \textbf{0.0153} & \textbf{0.0068} & \textbf{0.0029} \\
\midrule
\multicolumn{6}{c}{\textbf{Gender Recognition}} \\
\midrule
\multirow{2}{*}{HSJA \cite{chen2020hopskipjumpattack}} & SIM $\uparrow$ & 0.621 & 0.724 & 0.780 & 0.840 \\
 & LPIPS $\downarrow$ & 0.340 & 0.181 & 0.122 & 0.068 \\
\multirow{2}{*}{Latent-HSJA} & SIM $\uparrow$ & 0.794 & 0.817 & 0.821 & 0.824 \\
 & LPIPS $\downarrow$ & \textbf{0.052} & 0.041 & 0.039 & 0.037 \\
\multirow{2}{*}{LGC-H (Ours)} & SIM $\uparrow$ & \textbf{0.9974} & 0.9995 & \textbf{0.9997} & \textbf{0.9998} \\
 & LPIPS $\downarrow$ & 0.0950 & 0.0247 & 0.0159 & 0.0105 \\
\multirow{2}{*}{LGC (Ours)} & SIM $\uparrow$ & 0.9944 & \textbf{0.9996} & \textbf{0.9997} & \textbf{0.9998} \\
 & LPIPS $\downarrow$ & 0.1111 & \textbf{0.0217} & \textbf{0.0139} & \textbf{0.0097} \\
\bottomrule
\end{tabular}
}
\end{table}

\section{Experiments}
In this section, we conduct experiments to benchmark our proposed method, Latent Geometric Chords for Query-Efficient Decision-Based Attacks, LGC with its variant LGC-H, against current state-of-the-art adversarial attacks. Regardless of the target classifier or dataset, LGC and LGC-H consistently achieves superior results under both targeted and non-targeted threat models, while demonstrating a remarkable capability to preserve visual realism.

\subsection{Experimental Setting}

\textbf{Experiment hardware configuration:} Experiments are conducted using an Intel Xeon Platinum 8358P CPU and NVIDIA GeForce RTX 4090 GPU, running PyTorch 2.8.0 and Python 3.12.

\textbf{Datasets and Target Models:} We assess the effectiveness of LGC and LGC-H by using the ImageNet \cite{deng2009imagenet}, Places365 \cite{Zhou2018PlacesA1}, and CelebAMask-HQ \cite{lee2020maskgan} datasets. For the ImageNet dataset, we make the experiments against pre-trained VGG16 \cite{simonyan2014very}, ResNet-50 \cite{he2016deep}, DenseNet121 \cite{huang2017densely}, and Vision Transformer (ViT) \cite{dosovitskiy2020image} architectures which are implemented from the standard PyTorch library. For each target model, we randomly select 200 images for the non-targeted attack and 200 pairs of images for the targeted attack from the ILSVRC2012 validation set that are correclty classified by the target model. All input images are resized to $3 \times 224 \times 224$. For evaluations on the Places365 dataset, we consider ResNet-50 \cite{he2016deep} and DenseNet161 \cite{huang2017densely} provided directly by the MIT Places365 as target classifiers. Evaluations are similarly conducted using a randomly chosen 200 correctly classified images for non-targeted attacks and 200 pairs of correclty classified images for targeted attacks. For the CelebAMask-HQ \cite{lee2020maskgan} dataset, we consider a ResNet-18 \cite{he2016deep} model trained on the CelebAMask-HQ dataset for Identity and Gender Classification. Evaluations are conducted using a randomly chosen 100 correctly classified images for gender classification and 100 pairs of correctly classified images for targeted identity classification. 

\textbf{Baselines and Hyper-parameters:} We compare the performance of LGC and LGC-H with existing state-of-the-art attacks, specifically HSJA \cite{chen2020hopskipjumpattack}, CGBA \cite{reza2023cgba}, and Sign-OPT \cite{chen2020sign}. For the latent-space representation in LGC, we utilise a pre-trained VGG16-based Autoencoder \cite{horizon2333_imagenet_autoencoder} unless otherwise specified. Crucially, because the VGG16 latent space captures universal visual features and robust structural priors rather than domain-specific concepts, it generalises well beyond its training distribution. This allows us to use a single ImageNet-trained autoencoder to execute attacks on out-of-distribution datasets (e.g., Places365 and CelebAMask-HQ) without domain-specific retraining. For the CGBA baseline, we apply a frequency subspace reduction factor of $f = 4$. Furthermore, for both CGBA and our proposed LGC methods, we also set queires to estimate the initial normal vector as $N_0 = 30$ and the standard deviation for Gaussian sampling as $\sigma = 0.0002$ to estimate the normal vector. 

\textbf{Evaluation Metrics:} We employ the five metrics assessing both attack efficacy and visual stealth of LGC and LGC-H. The $L_2$ norm quantifies the mathematical magnitude of adversarial perturbations. Recognizing that mathematical distance often misaligns with human perception, we utilize the Structural Similarity Index Measure (SSIM) and Learned Perceptual Image Patch Similarity (LPIPS) to validate structural fidelity and visual imperceptibility. Moreover, attack effectiveness and computational efficiency are measured by analyzing the Attack Success Rate (ASR) against the number of model queries and thresholds of SSIM and LPIPS. For CelebAMask-HQ [40] dataset, we also use Structural Similarity (SIM).

\begin{figure*}[t]
    \centering
    \begin{minipage}[b]{0.46\textwidth}
        \centering
        \includegraphics[width=\linewidth]{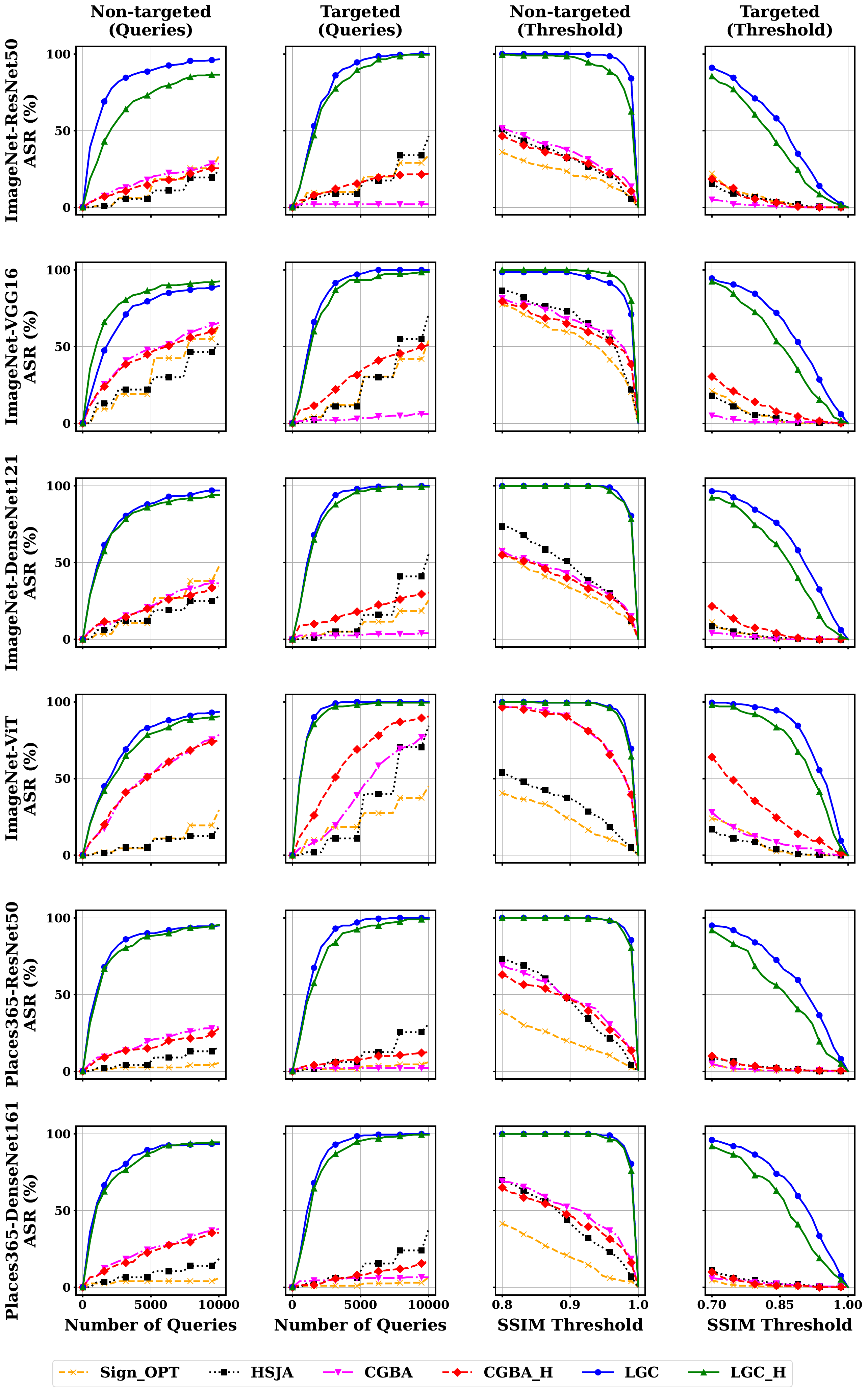}
        \caption{\scriptsize ASR versus queries and SSIM thresholds across various classifiers on ImageNet and Places365.}
        \label{fig:ssim_evaluation_final}
    \end{minipage}
    \hfill 
    \begin{minipage}[b]{0.46\textwidth}
        \centering
        \includegraphics[width=\linewidth]{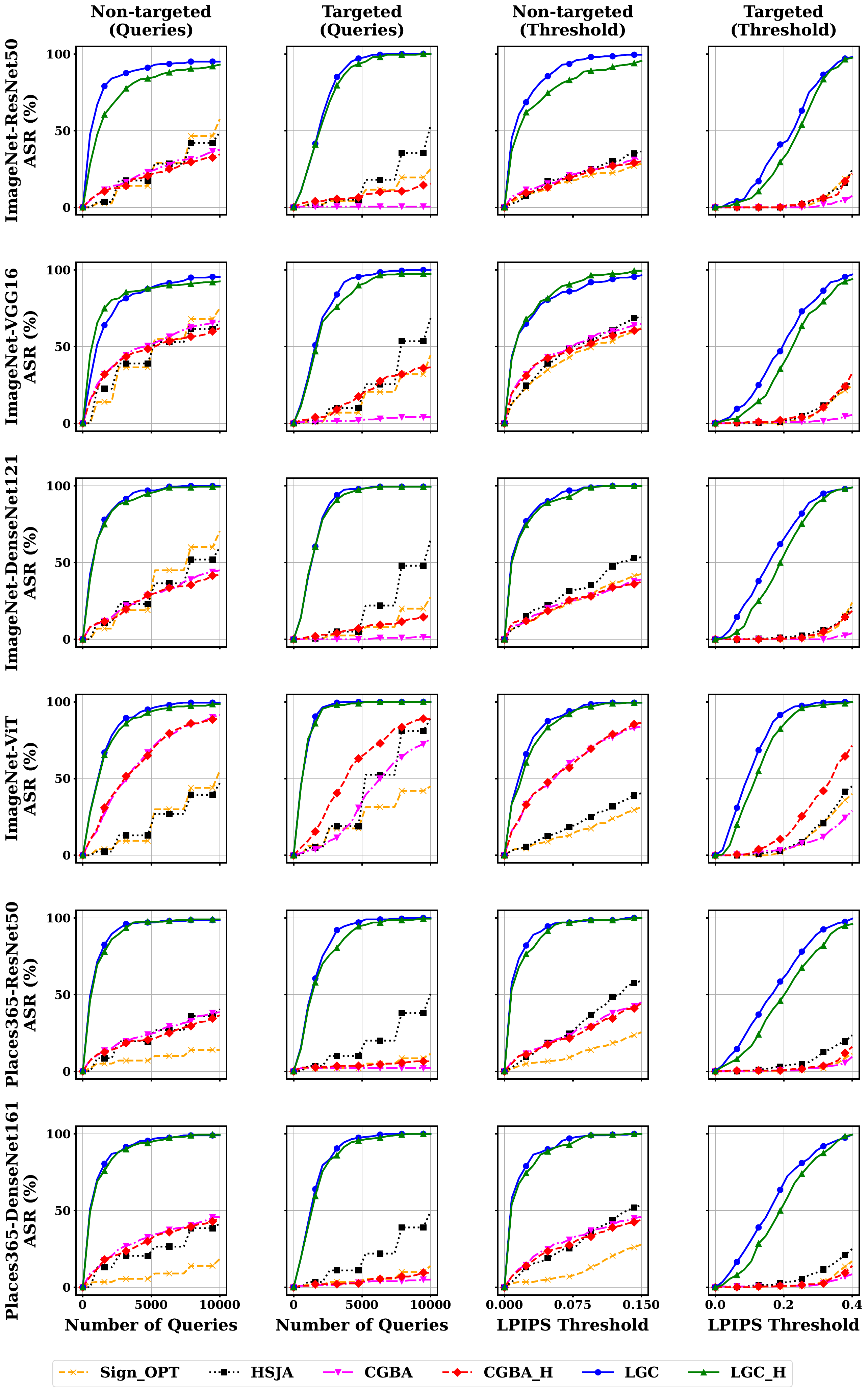}
        \caption{\scriptsize ASR versus queries and LPIPS thresholds across various classifiers on ImageNet and Places365.}
        \label{fig:lpips_evaluation_final}
    \end{minipage}
\end{figure*}

\begin{figure}[t]
    \raggedright 
    
    \subfloat[\parbox{0.5\textwidth}{\raggedright Granny Smith misclassified as arbitrary class (500 queries).} \label{fig:adv_500_1}]{
        \includegraphics[width=0.85\linewidth]{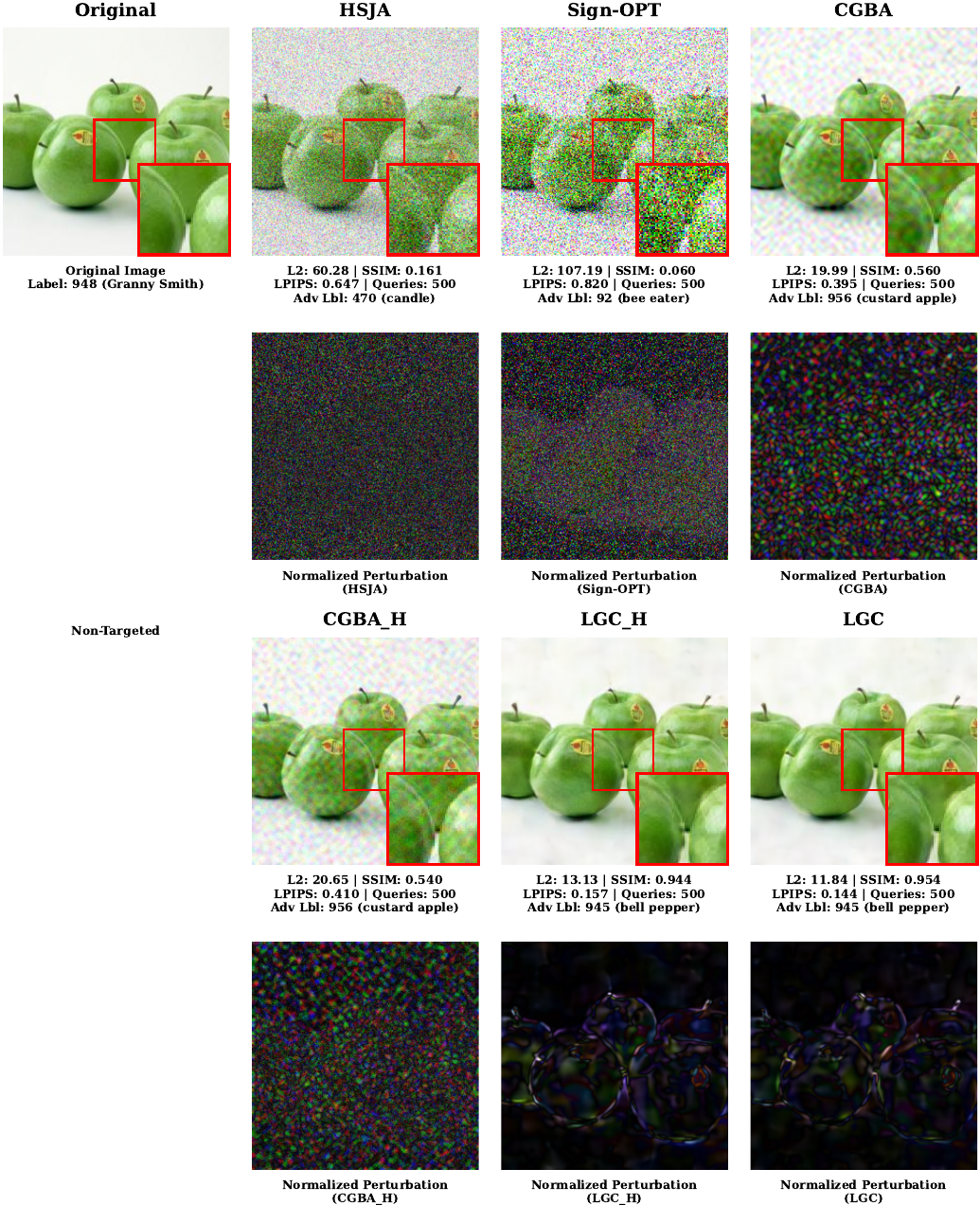}
    }
    
    \vspace{-3mm} 
    
    \subfloat[\parbox{0.5\textwidth}{\raggedright Barracouta misclassified as target class boathouse \\(5,000 queries).} \label{fig:targeted_5000_1}]{
        \includegraphics[width=0.85\linewidth]{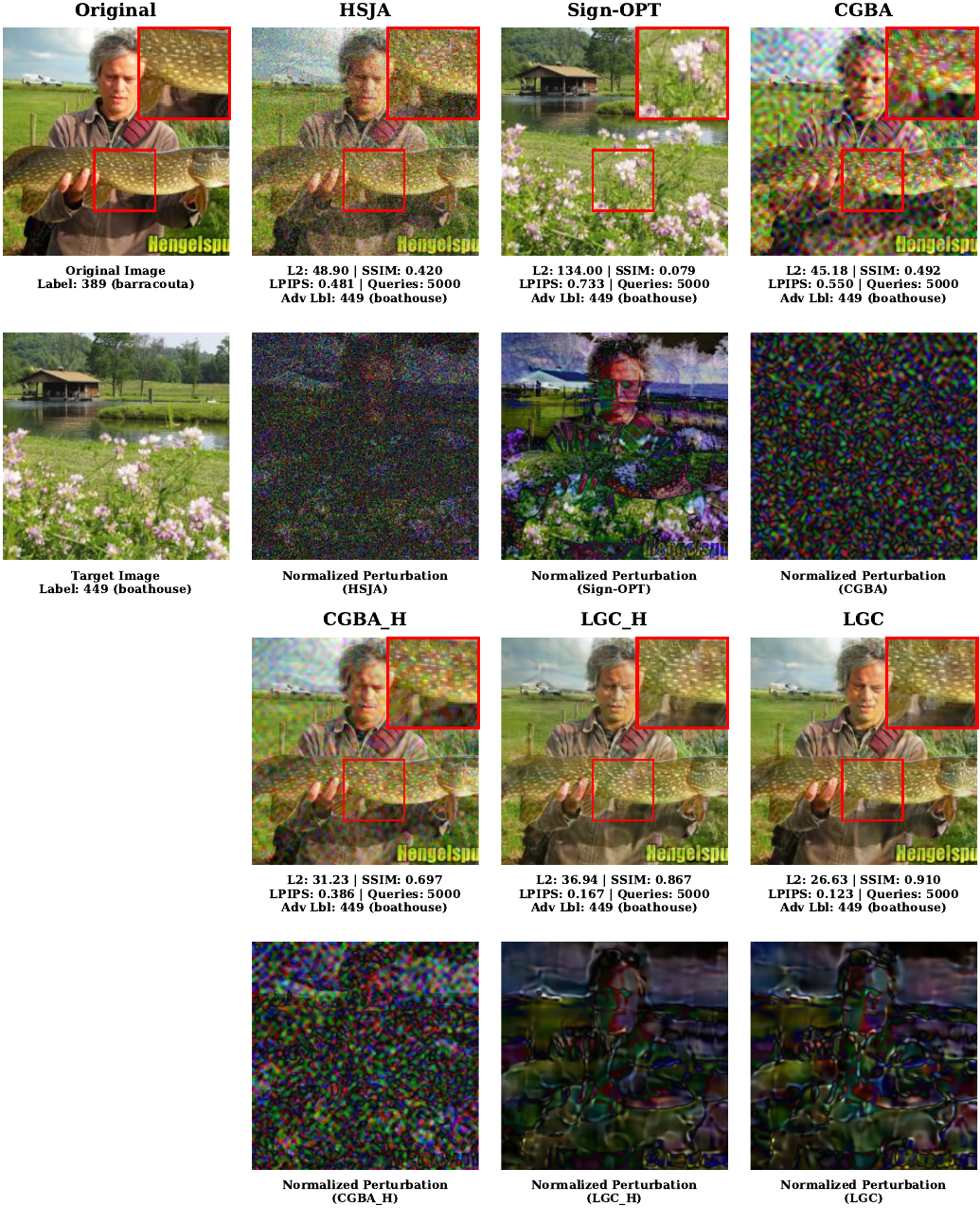}
    }
    \vspace{-4mm}
    \caption{\small Adversarial examples against ViT on ImageNet.}
    \label{fig:adversarial_comparison_combined_1}
\end{figure}

\begin{figure}[t]
    \raggedright

    \subfloat[\parbox{0.5\textwidth}{\raggedright Beach misclassified as arbitrary class (500 queries).} \label{fig:adv_500_2}]{
        \includegraphics[width=0.85\linewidth]{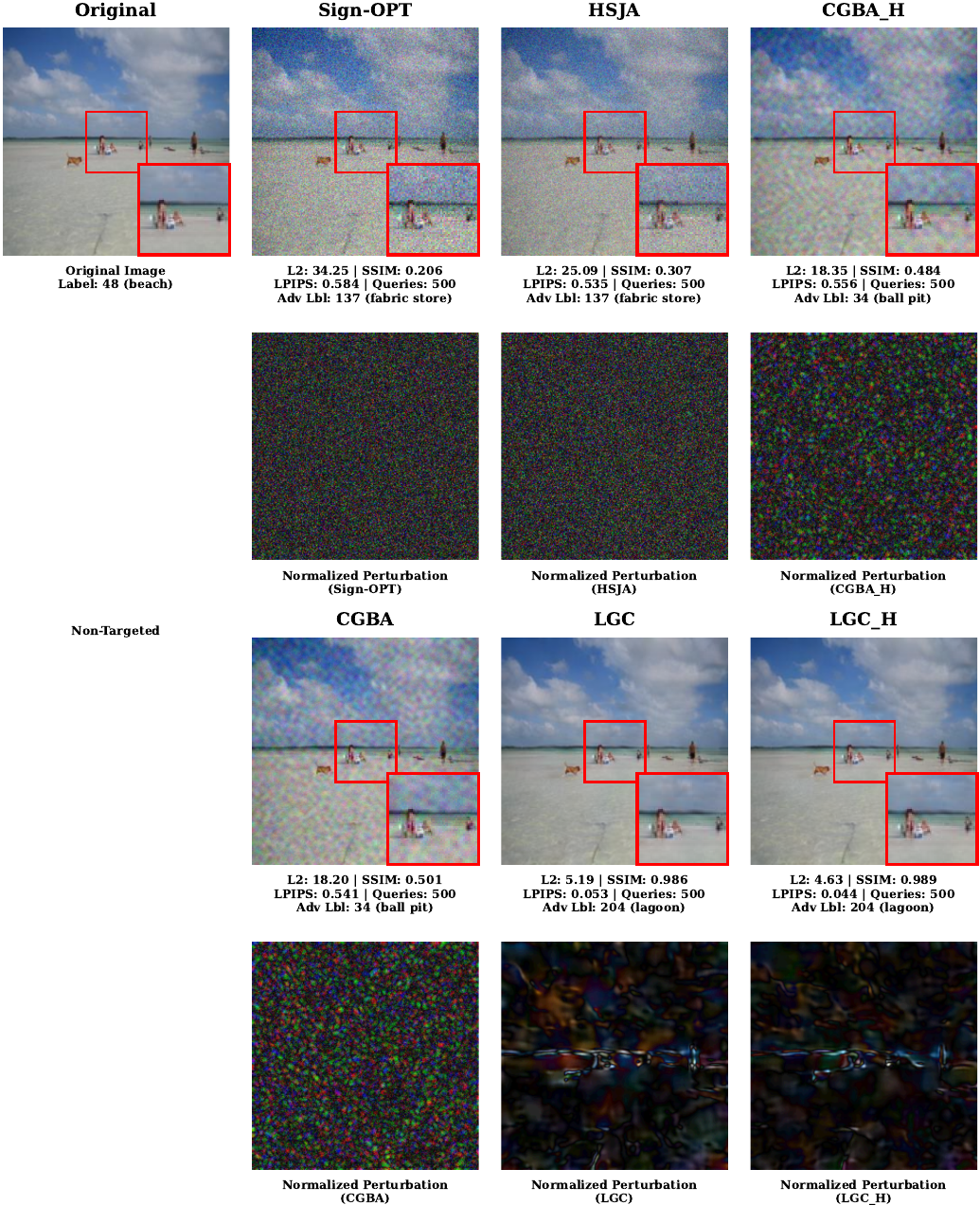}
    }
    
    \vspace{-3mm} 

    \subfloat[\parbox{0.5\textwidth}{\raggedright Attic misclassified as target class airport terminal \\(5,000 queries).} \label{fig:targeted_5000_2}]{
        \includegraphics[width=0.85\linewidth]{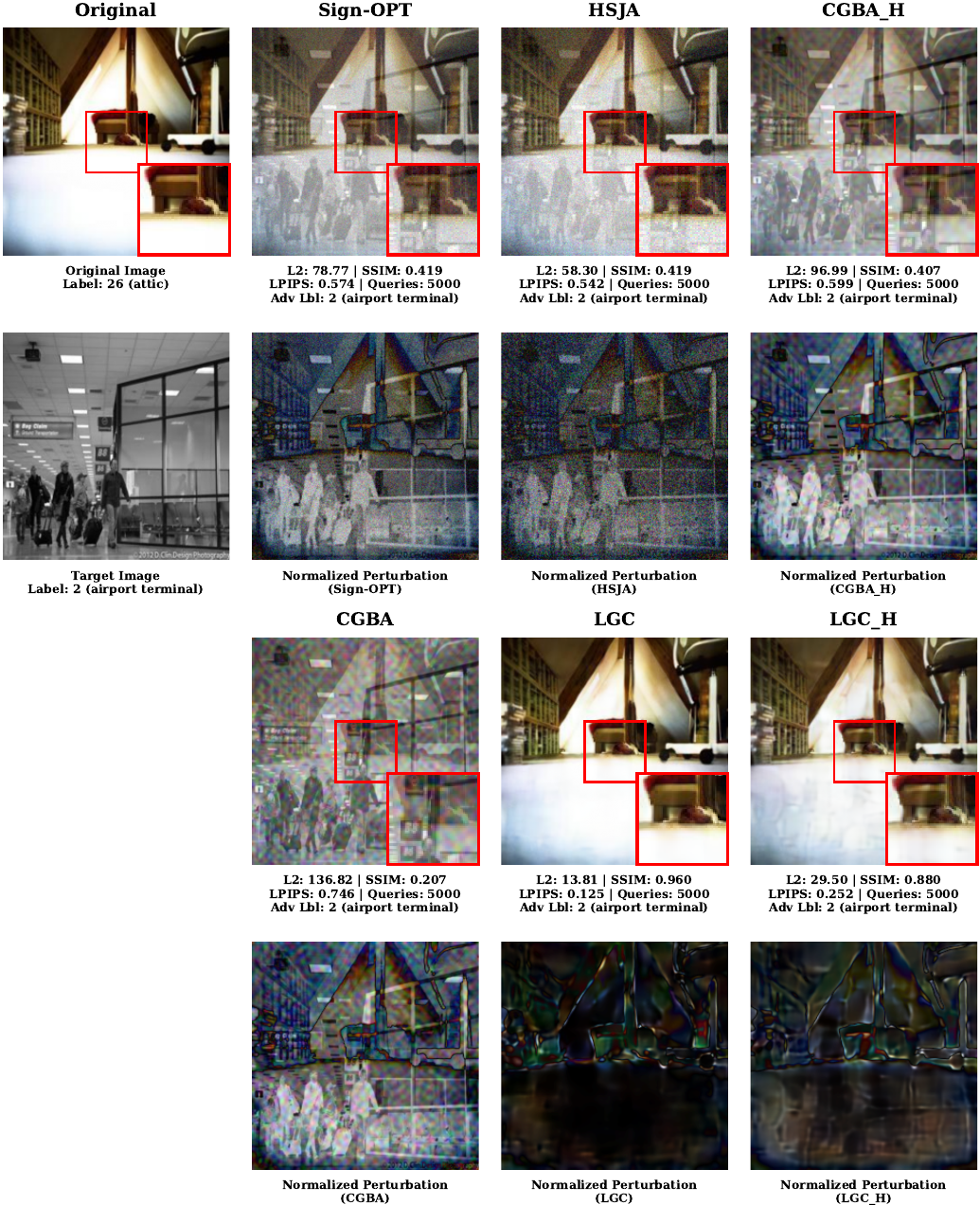}
    }
    \vspace{-4mm}
    \caption{\small Adversarial examples against ResNet-50 on Places365.}
    \label{fig:adversarial_comparison_combined_2}
\end{figure}

\begin{figure}[t]
    \raggedright 
    
    \subfloat[\parbox{0.5\textwidth}{\raggedright Female misclassified as class male.} \label{fig:adv_gender}]{
        \includegraphics[width=0.85\linewidth]{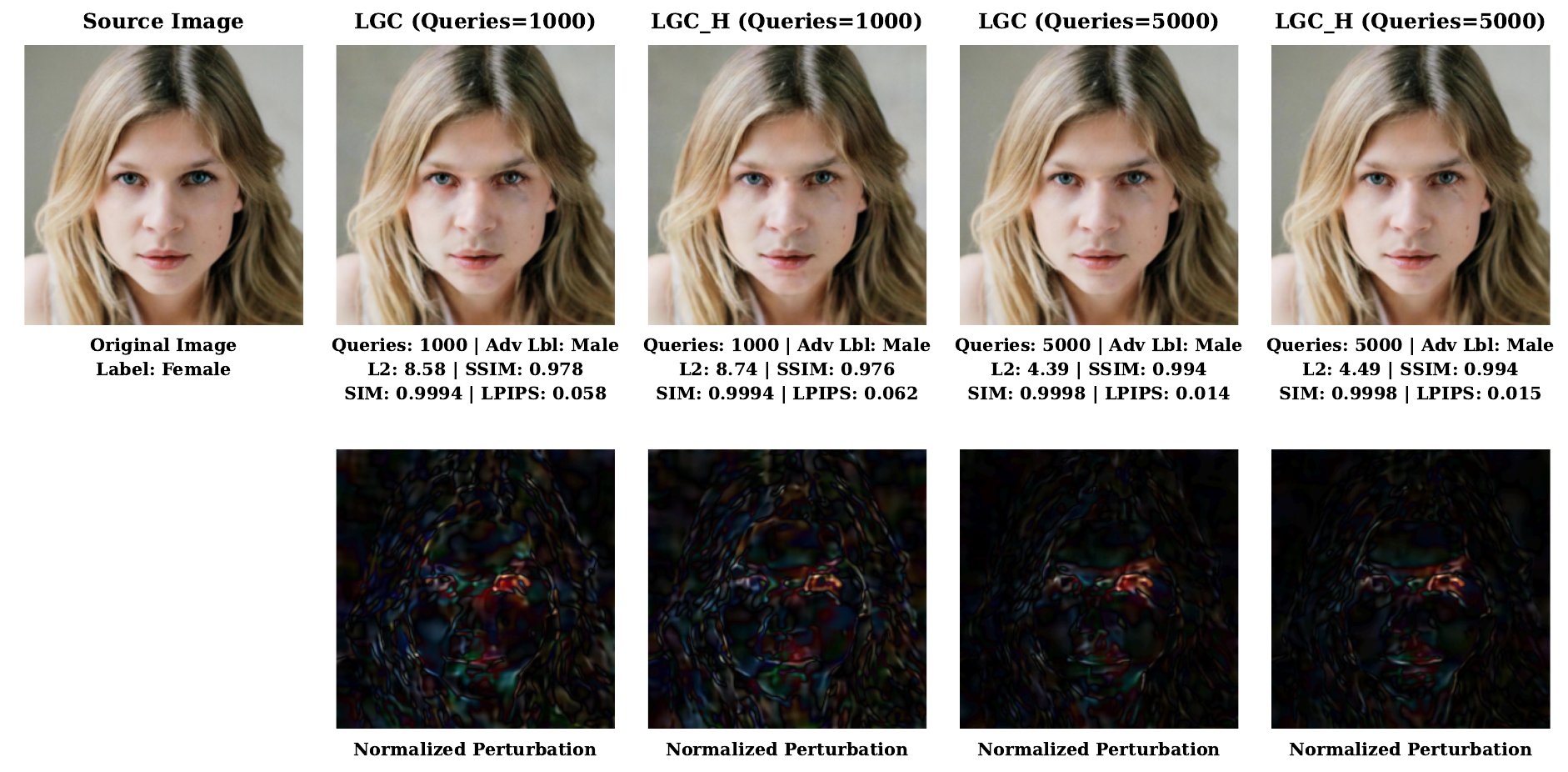}
    }
    
    \vspace{-3mm} 
    
    \subfloat[\parbox{0.5\textwidth}{\raggedright Class 255 misclassified as target class 124.} \label{fig:adv_targeted}]{
        \includegraphics[width=0.85\linewidth]{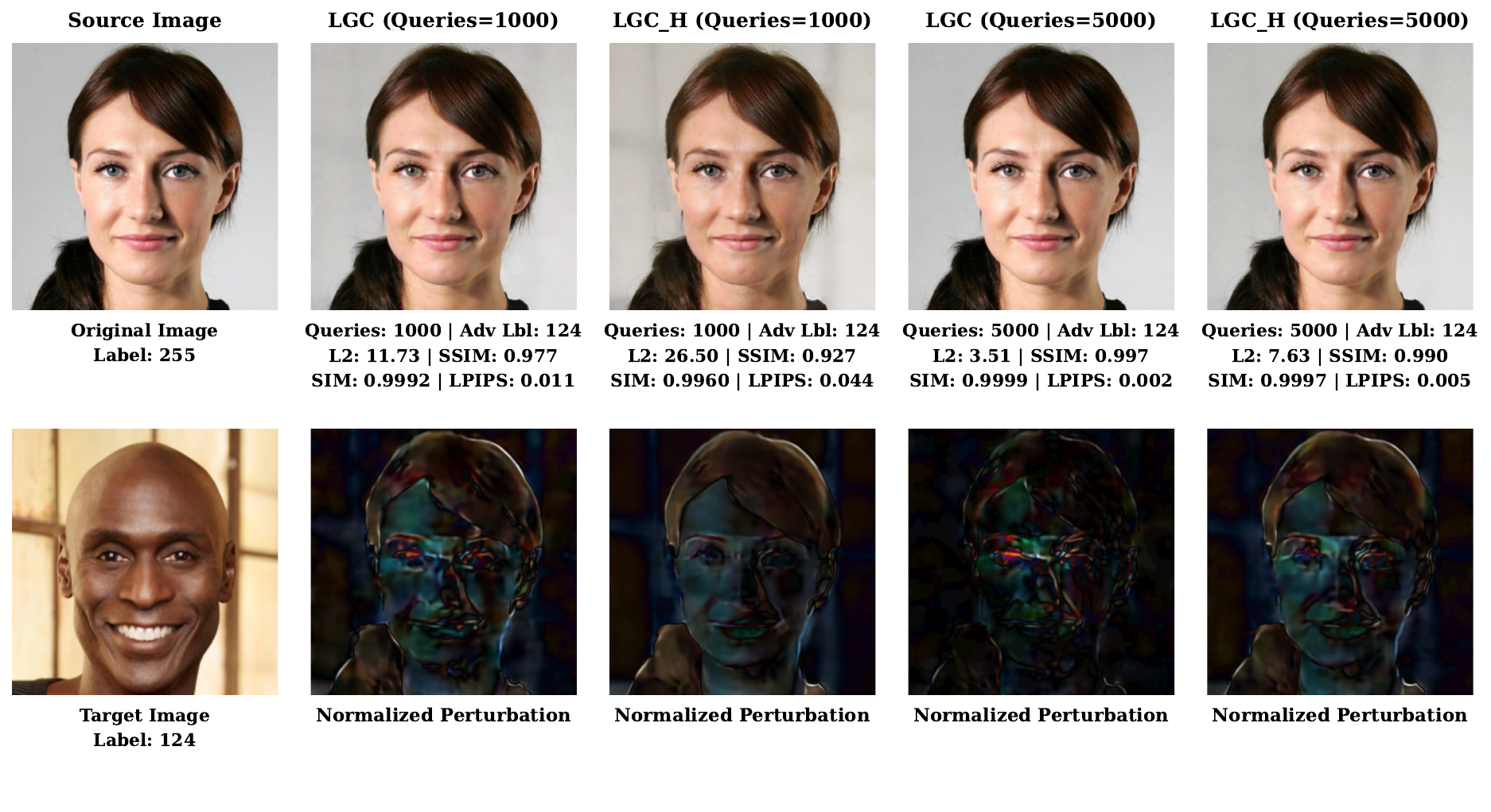}
    }
    \vspace{-4mm}
    \caption{\small Adversarial examples generated by the LGC and LGC-H methods against ResNet-18 on the CelebAMask-HQ dataset.}
    \label{fig:adversarial_comparison_combined}
\end{figure}

\begin{figure}[!t]
    \centering
    \includegraphics[width=0.90\columnwidth]{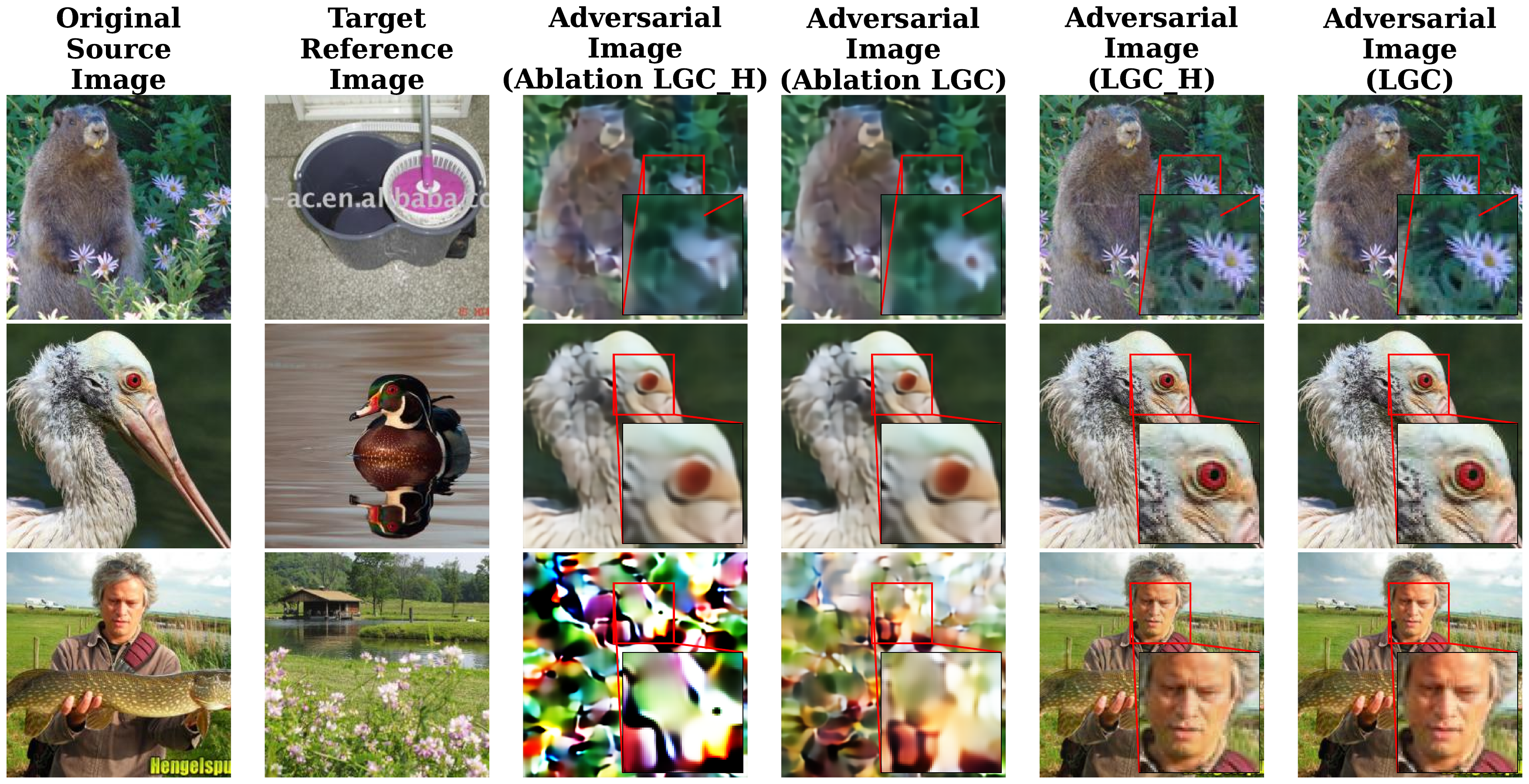}
    \caption{Ablation study comparing visual quality with and without Residual-based Adversarial Generation (RAG) at a query budget of $Q=10000$.}
    \label{fig:ablation_comparison}
\end{figure}

\begin{table*}[!t]
    \centering
    \scriptsize 
    \setlength{\tabcolsep}{5pt} 
    \renewcommand{\arraystretch}{0.85} 
    \setlength{\aboverulesep}{1.5pt} 
    \setlength{\belowrulesep}{1.5pt}
    
    \caption{Average (median) $L_{2}$ norm of perturbation for targeted and non-targeted black-box attacks under different query budgets against adversarially trained ViT on ImageNet dataset.}
    \label{tab:l2_at_results}
    \resizebox{\textwidth}{!}{
    \begin{tabular}{lll cccc cccc}
    \toprule
     &  &  & \multicolumn{4}{c}{Non-targeted} & \multicolumn{4}{c}{Targeted} \\[-0.001ex]
    \cmidrule(lr){4-7} \cmidrule(lr){8-11}
    Dataset & Model & Attack & 1000 & 2500 & 5000 & 10000 & 1000 & 2500 & 5000 & 10000 \\
    \midrule
    \multirow[c]{6}{*}{ImageNet} & \multirow[c]{6}{*}{ViT} & Sign\_OPT & 24.015(15.368) & 14.524(10.275) & 8.699(6.301) & 5.838(3.959) & 77.398(69.043) & 75.737(67.648) & 74.393(65.937) & 72.094(64.080) \\
 &  & HSJA & 23.112(18.000) & 17.519(13.130) & 13.914(10.981) & 11.196(8.714) & 65.655(57.382) & 60.680(52.237) & 53.809(45.727) & 44.681(36.064) \\
 &  & CGBA-H & 10.192(4.679) & 6.039(2.849) & 3.819(2.059) & \textbf{2.841(1.575)} & 54.159(45.912) & 42.785(33.436) & 32.868(22.184) & 24.611(12.986) \\
 &  & CGBA & 9.490(4.249) & 6.049(2.424) & 4.485(1.774) & 3.439(1.337) & 66.130(55.584) & 59.031(47.977) & 50.596(37.744) & 40.523(20.497) \\
 &  & LGC-H & \textbf{7.464(2.799)} & \textbf{4.715(2.162)} & \textbf{3.805(1.902)} & 3.379(1.799) & 54.847(46.482) & 42.277(35.091) & 31.830(26.128) & 23.104(16.767) \\
 &  & LGC & 7.915(2.640) & 5.299(2.051) & 3.817(1.805) & 3.297(1.656) & \textbf{49.846(41.854)} & \textbf{34.496(28.783)} & \textbf{24.125(18.304)} & \textbf{17.129(11.180)} \\
    \bottomrule
    \end{tabular}
    }

    \vspace{3mm}

    \caption{Average SSIM / LPIPS values of perturbation for targeted and non-targeted black-box attacks under different query budgets against adversarially trained ViT on ImageNet dataset.}
    \label{tab:ssim_lpips_at_results}
    \resizebox{\textwidth}{!}{
    \begin{tabular}{lll cccc cccc}
    \toprule
     &  &  & \multicolumn{4}{c}{Non-targeted} & \multicolumn{4}{c}{Targeted} \\[-0.001ex]
    \cmidrule(lr){4-7} \cmidrule(lr){8-11}
    Dataset & Model & Attack & 1000 & 2500 & 5000 & 10000 & 1000 & 2500 & 5000 & 10000 \\
    \midrule
    \multirow[c]{6}{*}{ImageNet} & \multirow[c]{6}{*}{ViT} & Sign\_OPT & 0.595 / 0.339 & 0.711 / 0.252 & 0.816 / 0.175 & 0.877 / 0.128 & 0.436 / 0.622 & 0.408 / 0.632 & 0.389 / 0.637 & 0.378 / 0.636 \\
     &  & HSJA & 0.574 / 0.349 & 0.660 / 0.285 & 0.721 / 0.240 & 0.775 / 0.201 & 0.434 / 0.572 & 0.385 / 0.579 & 0.362 / 0.572 & 0.394 / 0.526 \\
     &  & CGBA-H & 0.862 / 0.174 & 0.915 / 0.117 & 0.950 / 0.079 & 0.968 / 0.056 & 0.585 / 0.522 & 0.631 / 0.473 & 0.693 / 0.408 & 0.763 / 0.331 \\
     &  & CGBA & 0.881 / 0.154 & 0.931 / 0.098 & 0.953 / 0.070 & 0.969 / 0.050 & 0.411 / 0.636 & 0.409 / 0.620 & 0.460 / 0.572 & 0.560 / 0.484 \\
     &  & LGC-H & \textbf{0.969} / \textbf{0.046} & \textbf{0.987} / \textbf{0.021} & \textbf{0.993} / 0.013 & 0.995 / 0.008 & 0.794 / 0.299 & 0.847 / 0.233 & 0.889 / 0.174 & 0.923 / 0.122 \\
     &  & LGC & 0.966 / 0.048 & 0.984 / 0.023 & \textbf{0.993} / \textbf{0.012} & \textbf{0.996} / \textbf{0.007} & \textbf{0.809} / \textbf{0.284} & \textbf{0.872} / \textbf{0.203} & \textbf{0.914} / \textbf{0.140} & \textbf{0.941} / \textbf{0.094} \\
    \bottomrule
    \end{tabular}
    }
\end{table*}

\begin{figure}[!t]
    \raggedright 
    
    \subfloat[\parbox{0.5\textwidth}{\raggedright ASR versus queries and SSIM thresholds.} \label{fig:at_ssim_final}]{
        \includegraphics[width=0.85\linewidth]{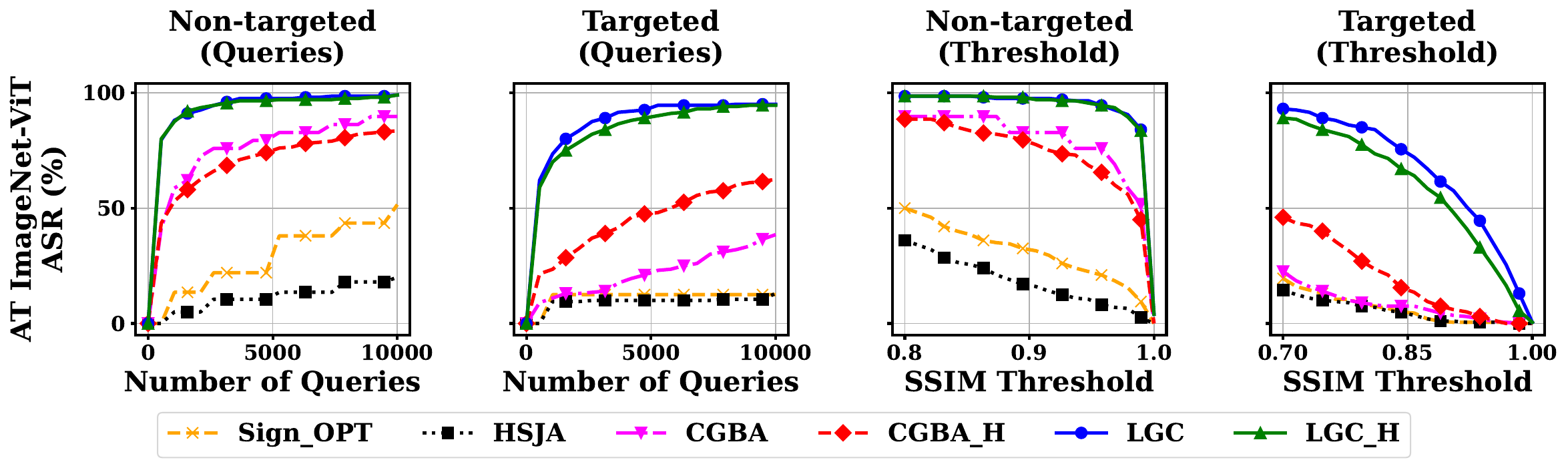}
    }
    
    \vspace{-3mm} 
    
    \subfloat[\parbox{0.5\textwidth}{\raggedright ASR versus queries and LPIPS thresholds.} \label{fig:at_lpips_final}]{
        \includegraphics[width=0.85\linewidth]{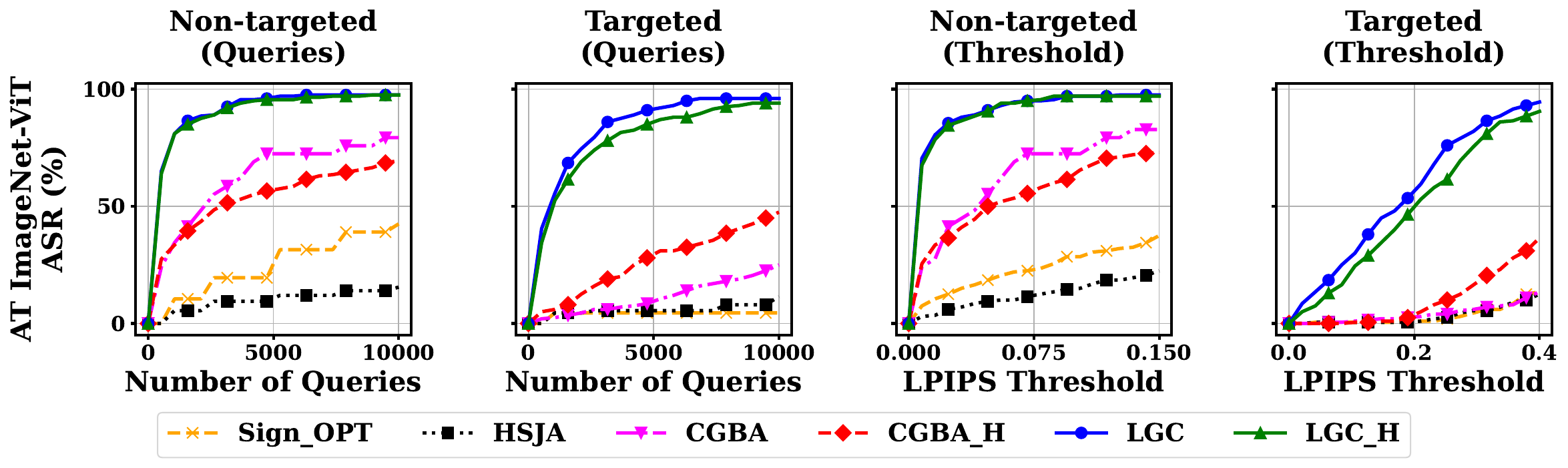}
    }
    \vspace{-4mm}
    \caption{\small ASR results  against Adversarially-trained ViT on ImageNet.}
    \label{fig:at_visual_quality}
\end{figure}

\begin{table*}[!t]
    \centering
    \scriptsize 
    \setlength{\tabcolsep}{5pt} 
    \renewcommand{\arraystretch}{0.85} 
    \setlength{\aboverulesep}{1.5pt} 
    \setlength{\belowrulesep}{1.5pt}

    \caption{Average (median) $L_{2}$ norm of perturbation for targeted and non-targeted black-box attacks under different query budgets against ViT on ImageNet dataset using different Autoencoders.}
    \label{tab:ae_l2_results}
    \resizebox{\textwidth}{!}{
    \begin{tabular}{lll cccc cccc}
    \toprule
     &  &  & \multicolumn{4}{c}{Non-targeted} & \multicolumn{4}{c}{Targeted} \\[-0.001ex]
    \cmidrule(lr){4-7} \cmidrule(lr){8-11}
    Dataset & Model & Attack & 1000 & 2500 & 5000 & 10000 & 1000 & 2500 & 5000 & 10000 \\
    \midrule
    \multirow[c]{4}{*}{ImageNet} & \multirow[c]{4}{*}{ViT} & LGC (VGG16 AE) & \textbf{11.041(8.850)} & \textbf{5.583(4.630)} & \textbf{3.885(3.454)} & \textbf{3.104(2.884)} & \textbf{44.215(42.285)} & \textbf{23.328(22.017)} & \textbf{11.965(10.747)} & \textbf{6.337(5.631)} \\
 &  & LGC-H (VGG16 AE) & 11.359(9.472) & 6.046(5.119) & 4.211(3.716) & 3.322(3.015) & 48.271(46.680) & 31.258(29.536) & 19.101(17.027) & 10.971(8.463) \\
 &  & LGC (ResNet-50 AE) & 11.294(9.052) & 5.925(4.728) & 4.079(3.545) & 3.229(3.007) & 52.875(44.156) & 31.563(22.065) & 18.598(10.412) & 10.851(6.181) \\
 &  & LGC-H (ResNet-50 AE) & 11.174(9.520) & 6.401(5.233) & 4.440(3.909) & 3.523(3.223) & 47.925(46.536) & 28.076(26.160) & 16.712(14.732) & 10.744(9.061) \\
    \bottomrule
    \end{tabular}
    }
    
    \vspace{3mm} 

    \caption{Average SSIM / LPIPS values of perturbation for targeted and non-targeted black-box attacks under different query budgets against ViT on ImageNet dataset using different Autoencoders}
    \label{tab:ae_ssim_lpips_results}
    \resizebox{\textwidth}{!}{
    \begin{tabular}{lll cccc cccc}
    \toprule
     &  &  & \multicolumn{4}{c}{Non-targeted} & \multicolumn{4}{c}{Targeted} \\[-0.001ex]
    \cmidrule(lr){4-7} \cmidrule(lr){8-11}
    Dataset & Model & Attack & 1000 & 2500 & 5000 & 10000 & 1000 & 2500 & 5000 & 10000 \\
    \midrule
    \multirow[c]{4}{*}{ImageNet} & \multirow[c]{4}{*}{ViT} & LGC (VGG16 AE) & \textbf{0.956} / \textbf{0.079} & \textbf{0.988} / \textbf{0.027} & \textbf{0.994} / \textbf{0.015} & \textbf{0.996} / \textbf{0.010} & \textbf{0.813} / \textbf{0.236} & \textbf{0.916} / \textbf{0.122} & \textbf{0.967} / \textbf{0.058} & \textbf{0.988} / \textbf{0.029} \\
     &  & LGC-H (VGG16 AE) & 0.955 / 0.081 & 0.986 / 0.031 & 0.993 / 0.017 & \textbf{0.996} / 0.012 & 0.804 / 0.241 & 0.888 / 0.146 & 0.942 / 0.081 & 0.973 / 0.043 \\
     &  & LGC (ResNet-50 AE) & 0.945 / 0.101 & 0.982 / 0.043 & 0.992 / 0.025 & 0.995 / 0.018 & 0.745 / 0.302 & 0.839 / 0.206 & 0.905 / 0.128 & 0.952 / 0.076 \\
     &  & LGC-H (ResNet-50 AE) & 0.946 / 0.100 & 0.980 / 0.048 & 0.991 / 0.029 & 0.994 / 0.021 & 0.802 / 0.249 & 0.897 / 0.150 & 0.950 / 0.087 & 0.976 / 0.055 \\
    \bottomrule
    \end{tabular}
    }
\end{table*}

\begin{figure}[!t]
    \raggedright

    \subfloat[\parbox{0.5\textwidth}{\raggedright ASR versus queries and SSIM thresholds.} \label{fig:ae_ssim_final}]{
        \includegraphics[width=0.9\linewidth]{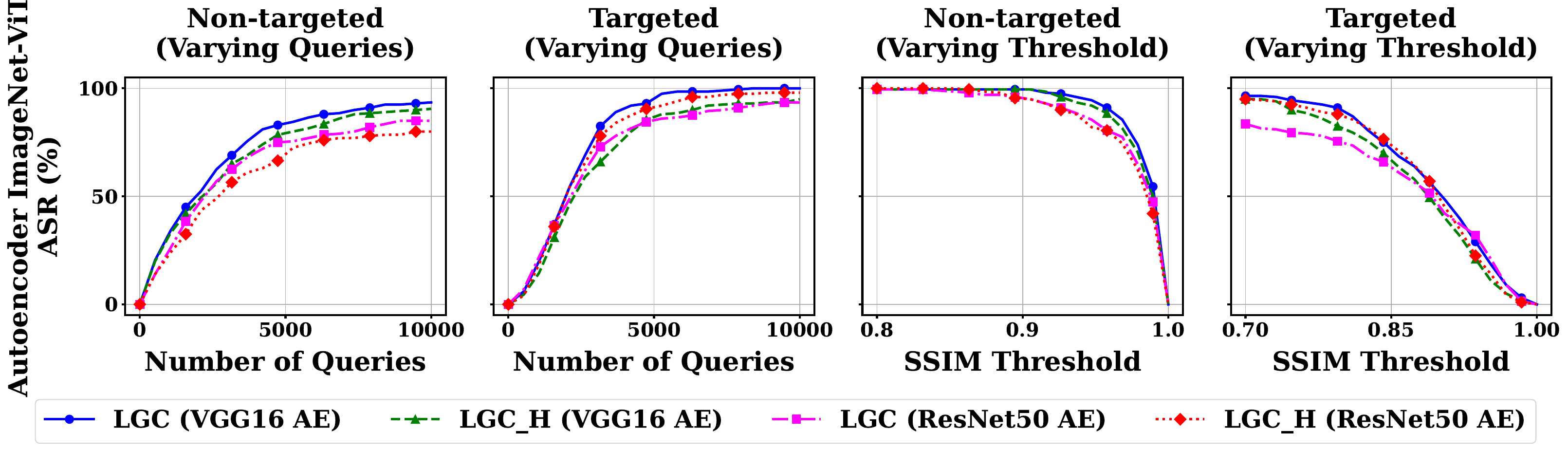}
    }
    
    \vspace{-3mm}
  
    \subfloat[\parbox{0.5\textwidth}{\raggedright ASR versus queries and LPIPS thresholds.} \label{fig:ae_lpips_final}]{
        \includegraphics[width=0.9\linewidth]{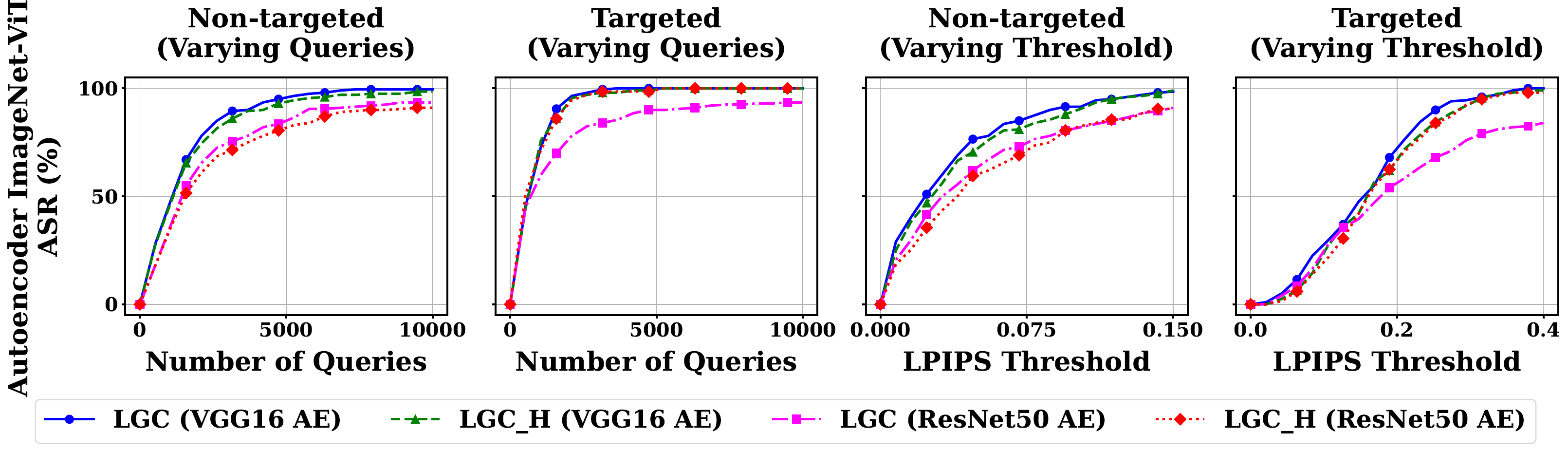}
    }
    \vspace{-4mm} 
    \caption{\small ASR results  against ViT varying autoencoder on ImageNet.}
    \label{fig:ae_visual_quality}
\end{figure}

\subsection{Experimental Results}

Conventional adversarial evaluations rely heavily on strict $L_2$ mathematical distances, a metric that often fails to reflect human visual perception. Effective adversarial stealth requires preserving natural image qualities. When evaluated on visual fidelity—specifically SSIM and LPIPS—our proposed LGC framework and its variant (LGC-H) consistently outperform all baseline methods on both the ImageNet and Places365 datasets. Table~\ref{tab:ssim_lpips_results} shows that LGC preserves high structural and perceptual quality across all classifiers. For instance, in the non-targeted ViT scenario, while CGBA-H mathematically minimizes the $L_2$ distance, its pixel-level noise causes noticeable structural degradation (SSIM 0.992, LPIPS 0.022). In contrast, LGC exhibits enhanced appearance (SSIM 0.996, LPIPS 0.010). This fidelity gap increases significantly in targeted attacks. For example, on ImageNet (ResNet-50), baselines like CGBA severely degrade the target image to a low SSIM of 0.326, whereas LGC preserves a high SSIM of 0.960 at 10,000 queries. This confirms that our semantic methodology effectively misleads classifiers while preserving the fundamental structure of the image.

To assess the practical effectiveness of existing attacks, we evaluate the Attack Success Rate (ASR) under strict perceptual stealth constraints. We plot ASR against the query budget using rigorous quality thresholds: SSIM $\geq$ 0.99 and LPIPS $\leq$ 0.05 for non-targeted scenarios, and SSIM $\geq$ 0.75 and LPIPS $\leq$ 0.3 for targeted scenarios. Additionally, we plot ASR against progressively stricter perceptual thresholds at a fixed budget of 3,000 queries, effectively penalizing methods that achieve misclassification by sacrificing visual realism.

As shown in Fig.~\ref{fig:ssim_evaluation_final}, under fixed SSIM constraints, LGC rapidly achieves high success rates. The fast convergence of the LGC curve reveals its ability to generate viable, high-fidelity adversarial examples within just 2,000 to 5,000 queries. This efficiency highlights a major limitation in current baselines: in targeted attacks under strict constraints, methods like CGBA and HSJA fail entirely, whereas LGC maintains an ASR of nearly 100\%. Furthermore, when evaluating against varying SSIM thresholds at a fixed budget, baseline performance drops sharply as the requirement approaches perfect structural fidelity (SSIM $\to$ 1.0). In contrast, LGC consistently sustains a high success rate under these strict constraints.

Similarly, Fig.~\ref{fig:lpips_evaluation_final} confirms this advantage using LPIPS evaluations. Under fixed LPIPS thresholds, LGC rapidly finds successful adversarial paths. When perceptual strictness increases (LPIPS $\to$ 0.0) at a fixed 3,000-query budget, baseline methods suffer severe performance drops, making them ineffective for stealthy deployments. LGC effectively overcomes these constraints, ensuring that the generated perturbations remain mathematically functional yet highly imperceptible to human observers.

Visual comparisons of the generated adversarial images and normalized perturbation maps at 1,000 and 10,000 queries are illustrated in Fig.~\ref{fig:adversarial_comparison_combined_1} and Fig.~\ref{fig:adversarial_comparison_combined_2}. Evaluated on the ViT (ImageNet) and ResNet-50 (Places365) models, these visualizations confirm that LGC eliminates unnatural, high-frequency noise during optimization. By effectively navigating the geometric decision boundary, LGC successfully circumvents the local minima that trap existing methods, yielding structurally preserved outcomes.

Finally, while prioritizing visual realism, LGC remains highly competitive in traditional $L_2$ norm minimization. As detailed in Table~\ref{tab:main_results}, LGC achieves state-of-the-art $L_2$ performance on Places365, reducing the average norm on ResNet-50 and DenseNet161 to 2.401 and 2.385 at 10,000 queries. Crucially, in highly challenging targeted scenarios, LGC significantly outperforms baselines by escaping local minima. On Places365 (ResNet-50) at 10,000 queries, LGC reduces the perturbation distance to 10.185, while CGBA stagnates at 83.097—an eight times smaller. This strong performance extends to ImageNet; against DenseNet121, LGC achieves an $L_2$ norm of 8.804 compared to CGBA's 77.448 (around a nine times decrease). In specific non-targeted evaluations (e.g., ViT and VGG16), LGC marginally trails baselines in raw $L_2$ magnitude (e.g., CGBA achieves 1.711 on ViT vs. LGC’s 3.104). However, as demonstrated above, accepting this small mathematical difference provides an optimal balance, preventing the severe structural damage caused by $L_2$-centric techniques.

To evaluate our framework, LGC and LGC-H, on structurally sensitive tasks, we attack a ResNet-18 model for Identity and Gender Classification using the CelebAMask-HQ dataset. Notably, autoencoders of LGC and LGC-H are pre-trained on ImageNet, underscoring its robust cross-domain transferability. Table~\ref{tab:combined_results} illustrates that LGC and LGC-H consistently outperform HSJA and Latent-HSJA baselines across all budgets. At 10,000 queries, LGC attains near-perfect Structural Similarity (SIM $\geq$ 0.9999) and ultra-low LPIPS (0.0029), substantially resolving the reconstruction bottlenecks that stall Latent-HSJA near 0.82. Visual results in Fig.~6 corroborate these gains; at 1,000 and 5,000 queries, both methods successfully deceive the classifier with minimal perturbations, making the adversarial examples perceptually identical to the source images.

\subsubsection{Performance against Adversarially Trained Models}

Defeating adversarially trained models represents a rigorous test of an attack's stealth and efficacy. We evaluated our framework against a state-of-the-art defended Vision Transformer, Mo2022When\_ViT-B, from the RobustBench library \cite{croce2021robustbench}. As shown in Table~\ref{tab:ssim_lpips_at_results}, LGC performs highly efficient early optimization in non-targeted settings, achieving high visual quality (SSIM up to 0.996, LPIPS 0.007). This visual advantage is clearly evident in targeted attacks. While baselines severely damage image structure to cross the robust boundary—dropping SSIM scores to 0.378 (Sign\_OPT), 0.394 (HSJA), and 0.560 (CGBA)—LGC maintains a high SSIM of 0.941 and an LPIPS of 0.094. Fig.~\ref{fig:at_visual_quality} shows that under strict visual constraints (SSIM $\geq$ 0.95/LPIPS $\leq$ 0.05 for non-targeted, and SSIM $\geq$ 0.75/LPIPS $\leq$ 0.3 for targeted), baseline success rates drop to near zero, whereas LGC succeeds reliably. Furthermore, LGC also outperforms baselines in mathematical metrics (Table~\ref{tab:l2_at_results}), reducing the targeted $L_2$ norm to 17.129 at 10,000 queries, compared to HSJA (44.681) and (Sign\_OPT) (72.049).

\subsubsection{Ablation Study on the RAG Mechanism}

To evaluate the impact of the Residual-based Adversarial Generation (RAG) mechanism, we conduct an ablation study, as shown in Fig. \ref{fig:ablation_comparison}.

When the RAG module is removed and the generative decoder output ($x_{final} = G(z)$) is used directly, the generated images exhibit significant blurring, color distortion, and loss of high-frequency details (see \textit{Ablation LGC-H} and \textit{Ablation LGC}). These artifacts highlight the inherent reconstruction errors of standard autoencoders.

Conversely, when the proposed RAG mechanism is applied (as formulated in Section IV-B), the isolated semantic shift is superimposed directly onto the pristine source image. As evidenced in the \textit{LGC-H} and \textit{LGC} columns, this preserves the original image's structural integrity. This visually demonstrates that RAG effectively circumvents decoder reconstruction errors, maintaining high visual fidelity that remains perceptually indistinguishable to human observers.

\subsubsection{Impact of Autoencoder Architecture on Semantic Perturbations}
To assess the influence of the autoencoder \cite{horizon2333_imagenet_autoencoder} backbone, we compared VGG16 against ResNet-50 on ImageNet using a ViT classifier. Tables~\ref{tab:ae_ssim_lpips_results} and \ref{tab:ae_l2_results} show that VGG16 consistently outperforms ResNet-50, maximizing structural preservation while minimizing the perturbation magnitude. Fig.~\ref{fig:ae_visual_quality} further validates that VGG16 achieves a higher ASR with significantly fewer queries under strict constraints (non-targeted: SSIM $\geq$ 0.99, LPIPS $\leq$ 0.05; targeted: SSIM $\geq$ 0.90, LPIPS $\leq$ 0.3). This advantage is rooted in VGG16’s strictly sequential architecture, which creates a feature space strongly aligned with human visual perception \cite{zhang2018unreasonable}. As a result, geometric operations within its latent space—such as LGC’s semicircular trajectories—translate predictably into visually coherent semantic shifts. This predictable mapping improves both search efficiency and visual fidelity. Conversely, ResNet utilizes identity skip connections ($H(x) = F(x) + x$) that distribute spatial representations across hierarchical scales \cite{he2016deep}. A localized step in ResNet's latent space simultaneously modifies features across these scales, mapping unpredictably to the pixel domain $\mathcal{X}$. Therefore, navigating ResNet's highly non-linear latent space requires substantially more queries compared to the sequentially aligned VGG backbone.

\section{Conclusion}

This study introduces a novel decision-based black-box framework: Latent Geometric Chords for Query-Efficient
Decision-Based Adversarial Attacks (LGC) with its variant, LGC-H. By executing a curvature-aware geometric search within a semantic manifold and generating adversarial images using a Residual-based Adversarial Generation method, LGC expands search dimensionality and significantly reduces generative reconstruction errors. Crucially, we establish a mathematical foundation proving that this chord-based formulation effectively expands the adversarial search space to a Hausdorff dimension of up to $2k$. This theoretical guarantee underpins the ability of LGC and LGC-H to efficiently bypass the strict dimensionality bottlenecks of standard manifold optimization without falling into suboptimal local minima. Evaluations on the ImageNet, Places365 and CelebAMask-HQ datasets demonstrate that LGC significantly outperforms existing baselines. In targeted attacks, it reduces the $L_2$ perturbation magnitude by a factor of six while preserving near-perfect visual fidelity (SSIM $> 0.96$, LPIPS $< 0.091$). Furthermore, LGC exhibits strong cross-dataset transferability and successfully compromises adversarially trained Vision Transformers. These results expose the limitations of current defenses against latent-derived semantic threats. Future work will investigate integrating Latent Geometric Chords into adversarial training methods and adapting this methodology for Latent Diffusion Models.

\end{document}